\documentclass{article}

\ifdefined\XeTeXrevision\else\ifdefined\pdfoutput\pdfoutput=1\fi\fi
\PassOptionsToPackage{sort}{natbib}
\usepackage[utf8]{inputenc}
\usepackage[T1]{fontenc}
\usepackage{afterquery}
\usepackage{graphicx}
\usepackage{amssymb}
\usepackage{booktabs}
\usepackage{longtable}
\usepackage{array}
\usepackage{float}
\usepackage[bottom]{footmisc}
\usepackage{url}
\usepackage{fvextra}
\DefineVerbatimEnvironment{promptverb}{Verbatim}%
  {breaklines=true,breaksymbolleft={},fontsize=\small,xleftmargin=0pt}
\usepackage{xcolor}
\usepackage{tikz}
\usepackage{tabularray}
\UseTblrLibrary{booktabs}
\DefTblrTemplate{contfoot-text}{blank}{}
\DefTblrTemplate{conthead-text}{blank}{}
\SetTblrTemplate{contfoot-text}{blank}
\SetTblrTemplate{conthead-text}{blank}
\usepackage{caption}
\newcommand{\swatch}[1]{%
  \begingroup\definecolor{modedisc}{HTML}{#1}%
  \tikz[baseline={([yshift=-0.5ex]s.center)}]{\node[rectangle, fill=modedisc, inner sep=0pt,
    minimum size=3.4ex] (s) {};}\endgroup}
\makeatletter
\newcommand{\modedisc}[2]{%
  \begingroup\definecolor{modedisc}{HTML}{#1}%
  \tikz[baseline={([yshift=\dp\@arstrutbox]d.south)}]{\node[circle, fill=modedisc, text=white, inner sep=0pt,
    minimum size=2.5em, font=\normalsize] (d) {#2};}\endgroup}
\makeatother
\newcommand{\modedot}[2]{%
  \begingroup\definecolor{modedisc}{HTML}{#1}%
  \raisebox{0pt}[0pt][0pt]{\tikz[baseline={([yshift=-0.3em]d.center)}]{%
    \node[circle, fill=modedisc, text=white, inner sep=0pt,
      minimum size=1em, font=\scriptsize] (d) {#2};}}\endgroup}
\usepackage{hyperref}
\newcommand{\figref}[2][]{\hyperref[#2]{Figure~\ref*{#2}#1}}
\newcommand{\tabref}[1]{\hyperref[#1]{Table~\ref*{#1}}}
\newcommand{\secref}[1]{\hyperref[#1]{Section~\ref*{#1}}}
\newcommand{\appref}[1]{\hyperref[#1]{Appendix~\ref*{#1}}}
\graphicspath{{figures/panels/}}

\title{OmicsBench: a long-horizon multi-turn benchmark of AI agents on research-study-scale bioinformatics}

\author{Gideon Shaked\aff{1,2}\corr, Siddharth Nagisetty\aff{1,2}}
\affiliations{\affnum{1}Department of Computational Medicine and Bioinformatics,
  University of Michigan \qquad \affnum{2}AfterQuery}
\corrnote{Corresponding author: \texttt{gshaked@umich.edu}}
\paperlinks{%
  \pnote{Code and data}{\href{https://huggingface.co/datasets/gideonshaked/OmicsBench}{\nolinkurl{huggingface.co/datasets/gideonshaked/OmicsBench}},\\\href{https://huggingface.co/datasets/gideonshaked/OmicsBench-grader-databases}{\nolinkurl{huggingface.co/datasets/gideonshaked/OmicsBench-grader-databases}}}%
}

\begin{document}

\maketitle

\begin{abstract}
Computational biologists increasingly delegate substantial analyses to AI agents in long-horizon, multi-turn sessions, yet most benchmarks of agents' bioinformatics capabilities pose a single short task with a single-shot prompt.
We present OmicsBench, a long-horizon, multi-turn bioinformatics benchmark for AI agents.
Each of its 97 tasks links dependent subtasks into a directed acyclic graph, for 662 underlying subtasks in all.
Each subtask is posed as one user turn and graded deterministically against an answer key, with optional in-context feedback on incorrect answers.
Across 12 models on three harnesses, the slowest agent averages 3.6 hours per task, and no combination comes near saturating the benchmark, with mean scores of 0.43 to 0.57 out of a possible 1 without in-context feedback and 0.54 to 0.67 with it.
Multi-turn evaluation also uncovers behavior that single-shot prompts cannot, as agents score lower on subtasks that depend on earlier subtasks, and in-context feedback helps agents recover some of that loss.
Because deterministic scoring gives no qualitative account of how an agent failed, we also introduce an unsupervised failure mode discovery pipeline that describes, embeds, and clusters agents' failed attempts to surface emergent failure modes.
Applied to OmicsBench, this pipeline exposes qualitative shortcomings in how agents carry out bioinformatics analyses, ranging from agents applying their own statistical filters in place of the specified ones to agents writing whole answers from memory.
\end{abstract}

\enlargethispage{\baselineskip}
\section{Introduction}
\label{sec:introduction}

AI agents are increasingly capable of carrying out computational biology tasks that require tool use, code execution, and reasoning over biological datasets \citep{alber_cellvoyager_2026,anthropic_evaluating_2026,nair_agentic_2026}.
Increasingly, computational biologists put these agents to work on long-horizon problems, in which raw data pass through many interdependent steps on the way to a scientific result, each building on the outputs of earlier ones.
Such dependencies form a directed acyclic graph (DAG) of analysis steps, the abstraction bioinformaticians already use to define pipelines in workflow systems, such as Snakemake and Nextflow \citep{koster_snakemakescalable_2012,di_tommaso_nextflow_2017}.
Typically, scientists remain in the loop, delegating a substantial analysis, reviewing what comes back and requesting revisions, and only then delegating the next.
Most bioinformatics benchmarks, by contrast, pose the whole analysis as a single prompt describing the desired output and grade only what the agent returns.
Such benchmarks therefore rarely measure how an agent performs over a realistic workflow, one dependent subtask at a time.

To address this gap, we present OmicsBench, a long-horizon, multi-turn benchmark of AI agents' computational biology capabilities (\figref{fig:protocol}).
Each of the 97 tasks in OmicsBench is a DAG of subtasks, 662 in all, that together make up a research-study-scale workflow, with each subtask graded against an answer key and each edge letting a subtask build on the results of earlier ones.
Every task is built on a real biological dataset, and its subtasks carry that dataset from raw data through to a scientific result.
The tasks are long-horizon, requiring agents to work for hours, with per-agent averages reaching 3.6 hours, 350 tool calls, and 110 million tokens per task.
OmicsBench poses each subtask as one user turn of a single session, either with no feedback on its answers (``single-pass mode'') or with in-context feedback on a failing answer and a chance to retry (``gated mode'').

Our contributions are:
\begin{itemize}
  \item \textbf{A long-horizon, multi-turn evaluation framework for bioinformatics.} OmicsBench poses a research-study-scale analysis as a DAG of dependent subtasks.
    To our knowledge, it is the first bioinformatics benchmark to pose subtasks as the turns of one session and to give in-context feedback.
    The framework is domain-agnostic, so any research study that is expressible as a DAG of deterministically gradable steps can be implemented as an OmicsBench task, and the benchmark scales to DAGs of arbitrary length and complexity.
  \item \textbf{A benchmark of 97 research-study-scale tasks and 662 subtasks, and an evaluation of 12 frontier models on three harnesses.} Built from published studies and their datasets, OmicsBench uses 58 biologically informed graders that match named entities, such as genes and organisms, via database records rather than matching their names as strings.
    We release the runner, all graders, and three tasks with their datasets, while the remaining 94 tasks form a private test set.
    No model/harness combination comes close to saturating the benchmark.
    We find that agents' scores decline on later turns and dependent subtasks, while in-context feedback helps them recover part of that loss.
  \item \textbf{An unsupervised pipeline for discovering agents' failure modes.} The pipeline describes, embeds, and clusters failed attempts.
    It surfaces 12 emergent, bioinformatics-specific failure modes beyond those of a priori annotation, ranging from agents applying their own selection criteria to agents fabricating answers from memory.
    These modes identify concrete behaviors for future models to improve upon and provide valuable post-training signals.
\end{itemize}

\section{Related work}
\label{sec:related-work}

OmicsBench sits between two lines of prior work, short-horizon benchmarks that ask for one small piece of analysis and long-horizon benchmarks that ask for work on the scale of a whole research study, and OmicsBench combines the scale of the latter with the granularity of the former.

\subsection{Short-horizon bioinformatics benchmarks}
\label{sec:related-short}

These benchmarks evaluate self-contained pieces of work whose results are not carried into later requests.
Some, including LAB-Bench \citep{laurent_lab-bench_2024}, LABBench2 \citep{laurent_labbench2_2026}, Genome-Bench \citep{yin_toward_2025}, BioProBench \citep{liu_bioprobench_2026}, BioCoder \citep{tang_biocoder_2024}, BioDSBench \citep{wang_can_2025}, and CellVoyager's CellBench \citep{alber_cellvoyager_2026}, test capabilities such as biological knowledge, protocol reasoning, analysis planning, and code generation, without an agent working with a supplied dataset in a coding environment.
Others ask an agent to analyze data in a coding environment and answer a question in one request.
These include GenoTEX \citep{liu_genotex_2025}, BAISBench \citep{luo_benchmarking_2026}, BiomniBench-DA \citep{qu_biomnibench_2026}, CompBioBench \citep{nair_agentic_2026}, BioMysteryBench \citep{anthropic_evaluating_2026}, DrugDiscoveryBench \citep{akyurek_drugdiscoverybench_2026}, and LatchBio's SpatialBench \citep{workman_spatialbench_2026}, scBench \citep{workman_scbench_2026}, EpiBench \citep{muralidharan_epibench_2026}, VariantBench \citep{bhowmick_variantbench_2026}, TxBench-PP \citep{le_txbench-pp_2026}, TxBench-Antibody Discovery \citep{ramasamy_txbench-antibody_2026}, and TxBench-Oligonucleotide Discovery \citep{jacko_txbench-oligonucleotide_2026}.
Because each item ends after one request, these benchmarks do not measure whether an agent can carry intermediate results through a research-study-scale sequence of dependent subtasks.

\subsection{Long-horizon bioinformatics benchmarks}
\label{sec:related-long}

These benchmarks evaluate research-study-scale analyses, but they present the entire objective in a single request.
BixBench \citep{mitchener_bixbench_2025}, BixBench3 \citep{koch_bixbench3_2026}, GeneBench \citep{li_genebench_2026}, GeneBench-Pro \citep{li_genebench-pro_2026}, BioAgent Bench \citep{fa_bioagent_2026}, BioML-bench \citep{miller_bioml-bench_2025}, SpatialBench-Long \citep{diks_verifiable_2026}, scBench-Long \citep{diks_scbench-long_2026}, and the life-science tasks in Terminal-Bench-Science \citep{terminal-bench-science_team_terminal-bench-science_2026} give an agent data and a research objective, allow it to work without further instruction, and grade its final output against a published result, a reproduced or deterministic ground truth, or an expert rubric.
Some of these benchmarks already structure their tasks around dependent steps, such as BixBench3's artifact dependency graph and the cascaded decision points in GeneBench and GeneBench-Pro.
Those dependent steps nevertheless remain inside a single prompt, so the benchmarks neither present successive subtasks as turns in one session nor provide in-context feedback on one step before the next depends on it.
Separately, benchmarks that characterize agent failures generally assign attempts to predefined categories, leaving emergent failure modes unobserved.

\section{OmicsBench}
\label{sec:omicsbench}

\subsection{OmicsBench formulation}
\label{sec:formulation}

Each OmicsBench task $T = (P, D, G)$ is composed of a task prompt $P$, a biological dataset $D$, and a task DAG $G = (V, E)$, in which each node $v \in V$ is a subtask and each edge $e = (u, v) \in E$ means subtask $v$ depends on the answer to subtask $u$ (\figref[A]{fig:protocol}).
Each subtask $v = (v_p, v_e)$ consists of a subtask prompt $v_p$ and an evaluation set $v_e$ (\tabref{tab:notation}).
The evaluation set holds one or more deterministic graders $g_i$, each paired with an answer key $y_i$, so that $v_e = \{(g_1, y_1), (g_2, y_2), \ldots\}$.
An agent answers the subtask prompt by writing an answer file $a_v$.
Each grader in $v_e$ is assigned one field of the answer file $a_v$ and scores that field against its answer key, giving one score per grader, $g_i(a_v, y_i) \in [0, 1]$.
A subtask's score $s_v$ is the mean of these grader scores, and the task score $S$ is the mean of all subtask scores,
\[s_v = \frac{1}{|v_e|} \sum_i g_i(a_v, y_i), \qquad S = \frac{1}{|V|} \sum_{v \in V} s_v.\]

To evaluate an agent on a task, we open the session with the benchmark system prompt and the task prompt, which together describe the problem and the dataset (\appref{app:system-prompt}), and the task DAG is then unrolled in dependency order, with every subtask prompt becoming a user turn in that same continuous session (\appref{app:prompt-task}).
Because the session is continuous, the agent's previous answer files stay on hand in the execution container, and its reasoning for those answers stays in its context.
A turn ends when the agent stops working or when a time limit $t_{\max}$ is reached, whichever comes first, and whatever answer file is on disk at that moment is the one graded.

In single-pass mode, the agent receives no feedback on its answers, and when it stops working, the next subtask prompt follows (\figref[B]{fig:protocol}).
In gated mode, an answer scoring below a gate threshold $\tau$ is returned to the agent as a failure, and it is re-prompted to fix its work (\appref{app:prompt-retry}) until an answer scores at or above $\tau$, or it reaches an attempt limit $\theta$ (\figref[C]{fig:protocol}, \tabref{tab:parameters}).
In either mode, a subtask's recorded score is that of its best attempt, and when the best attempt was not the last, the workspace is restored to it before the next subtask is posed (\appref{app:prompt-task}).
Single-pass mode measures what an agent can produce unaided, and gated mode measures how much in-context feedback allows it to recover its score (\figref[B]{fig:pipeline}).

\subsection{Benchmark construction}
\label{sec:benchmark-construction}

OmicsBench tasks are built from published papers and their accompanying biological datasets by a pipeline of agents and of human experts with graduate-level computational biology training (\figref[A]{fig:pipeline}).
Human experts first curate papers, whose reported results serve as ground truth, and their accompanying biological datasets.
A writer agent then uses the papers and datasets to draft OmicsBench tasks, as in BixBench3 \citep{koch_bixbench3_2026}.
A reviewer agent confirms that the tasks are supported by their papers and free of data leakage, or sends them back to the writer, rejecting any task that is unfixable or still flawed after three rounds of revision.
Automated quality checks confirm that each task is in valid OmicsBench format, and human experts then review and edit each task, or reject it.
Three of the 97 tasks are publicly released, and the remaining 94 are held back as a private test set to keep the benchmark useful for evaluating future models (\appref{app:code-data-availability}).

The task datasets span a wide range of data modalities, with transcriptomic assays in 66 tasks, proteomic assays in 29, and epigenomic assays in 9, with 21 tasks drawing on more than one modality (\figref[A]{fig:composition}).
The datasets come from ten different organisms (\figref[D]{fig:composition}), have a median of 3 files and a maximum of 32 (\figref[G]{fig:composition}), and range in size from 365 KB to 8.4 GB (\figref[H]{fig:composition}).
The biology that the tasks address covers 11 biological domains (\figref[B]{fig:composition}), and the subtasks cover 18 common bioinformatics analysis types (\figref[C]{fig:composition}).
Most tasks have between 4 and 10 subtasks, with a median of 7 (\figref[E]{fig:composition}), and in the median task the longest chain of dependent subtasks runs to 3 (\figref[F]{fig:composition}), with many tasks featuring branching and joining structures in their DAGs.

\begin{figure}[H]
\centering
\includegraphics[width=\textwidth]{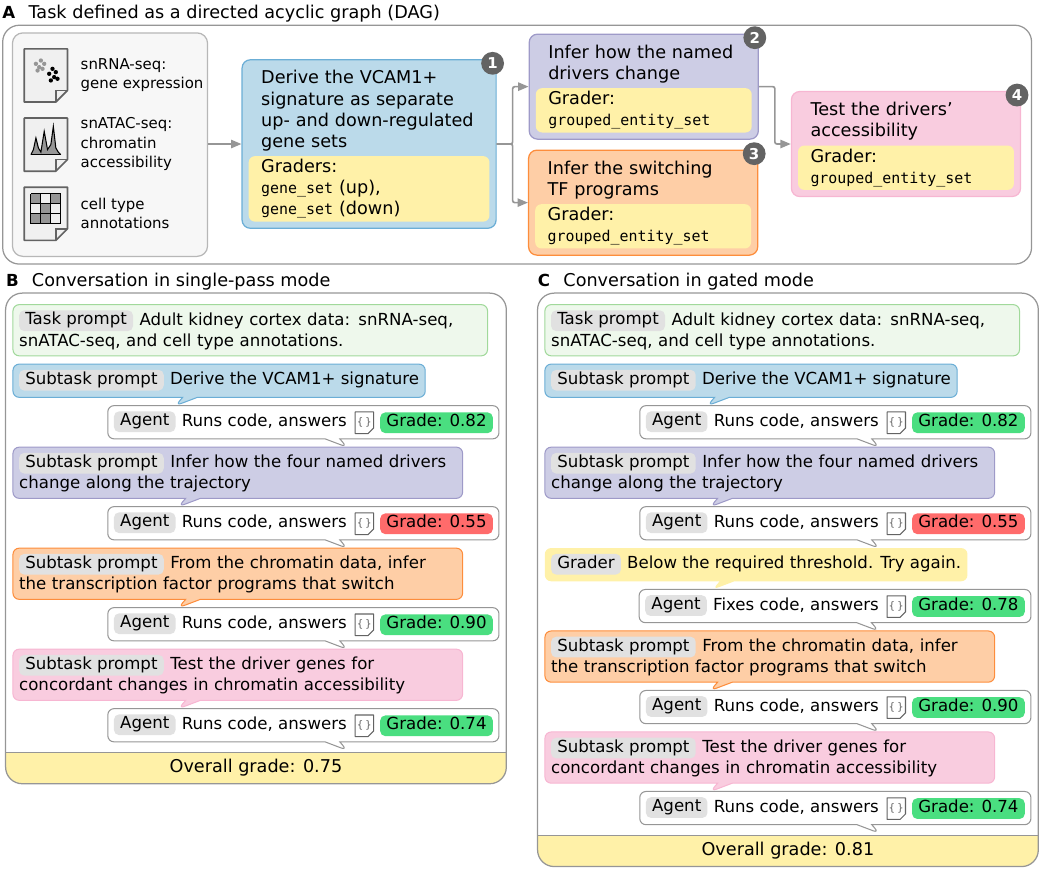}
\caption{\textbf{Example OmicsBench task.} A stylized view of the example task in \appref{app:example-question} (\tabref{tab:example-data}, \tabref{tab:example-tasks}). (\textbf{A}) A task is a DAG of analysis subtasks on a dataset. (\textbf{B}) Single-pass mode, with each subtask one user turn and no feedback on answers. (\textbf{C}) Gated mode, where an answer scoring below the gate threshold ($\tau$) is returned for a retry in the same session, until an answer reaches $\tau$ or an attempt limit.}
\label{fig:protocol}
\end{figure}

\begin{figure}[!b]
\centering
\includegraphics[width=\textwidth]{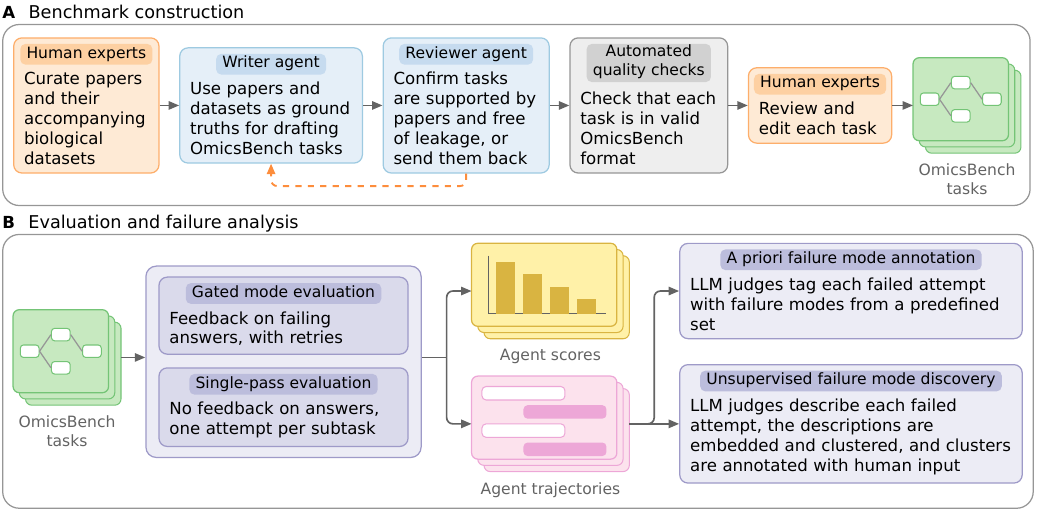}
\caption{\textbf{Construction and evaluation of OmicsBench.} (\textbf{A}) Benchmark construction. (\textbf{B}) Evaluation and failure analysis pipeline.}
\label{fig:pipeline}
\end{figure}

\subsection{Grading}
\label{sec:grading}

Because biological results take varied forms and rarely reduce to single values compared for equality, OmicsBench uses biologically informed graders, each built for one kind of answer (\tabref{tab:grader-catalog}).
For example, two correct enrichment analyses may order their terms differently and report slightly different p-values, so the \texttt{enrichment\_result\_set} grader requires the terms to match, compares p-values on a log scale, and ignores order.
Each of the other 57 graders likewise checks what is biologically meaningful in its kind of answer and ignores the rest, with equivalent behavior for biological data types ranging from dose-response curves to genomic intervals to docking poses.

Furthermore, many types of biological entities are referred to by multiple identifiers.
For example, the gene CD274 is also known as PD-L1 and B7-H1, and a string match would mark any of the three valid identifiers wrong against an answer key that uses another.
OmicsBench graders therefore match genes, organisms, metabolites, microRNAs, and antimicrobial resistance genes through a reference database for each kind of entity rather than by exact string matching (\tabref{tab:identifier-databases}).

\subsection{Failure mode analysis}
\label{sec:failure-mode-analysis}

We characterized failed attempts with both a priori failure mode annotation and an unsupervised failure mode discovery pipeline that surfaces emergent failure modes in agents' biological reasoning (\figref[B]{fig:pipeline}).
For both, we drew a single sample of 150 of the 662 subtasks, stratified by turn position and by the kind of result the subtask asks for, and analyzed every attempt scoring below $\tau$ at those subtasks.
We used only single-pass trajectories, so that every failed attempt is an agent's first and only try at a subtask.
All LLM judges were Opus 5 at temperature 0 (\appref{app:judge-prompts}).

Our a priori vocabulary of 13 failure modes is composed of the ten defined in the BixBench3 study \citep{koch_bixbench3_2026} as well as three we added for answer sets that are too broad, too strict, or selected on the wrong criterion (\tabref{tab:apriori-modes}).
For each failed attempt, an LLM judge wrote a short narrative of the approach the agent took and how that approach produced the submitted answer, and assigned as many failure modes from the vocabulary as the evidence supported (\appref{app:prompt-annotate}).
Repeated annotation passes of the LLM judge agreed with a Cohen's kappa of 0.76 (\appref{app:apriori-modes}).

The unsupervised failure mode discovery pipeline that we built describes, embeds, and clusters failed attempts to group those that failed in the same way, surfacing emergent failure modes that a predefined set would miss.
An LLM judge writes a short narrative of each failed attempt and rewrites it as a task-agnostic description of how the submitted answer diverged from the answer key (\appref{app:prompt-abstract}).
The pipeline then embeds the descriptions, reduces them with UMAP \citep{mcinnes_umap_2018}, and clusters them with HDBSCAN \citep{campello_density-based_2013}, which groups only the points that lie in dense regions and leaves the rest unassigned, so it does not force every attempt into a cluster.
A separate LLM judge writes a name and definition for each cluster from the descriptions nearest its center (\appref{app:prompt-name}).
We edited every name and definition by hand for readability and grouped the failure modes into five families by the mechanism of failure they share (\appref{app:emergent-modes}).

\begin{figure}[H]
\centering
\includegraphics[width=\textwidth]{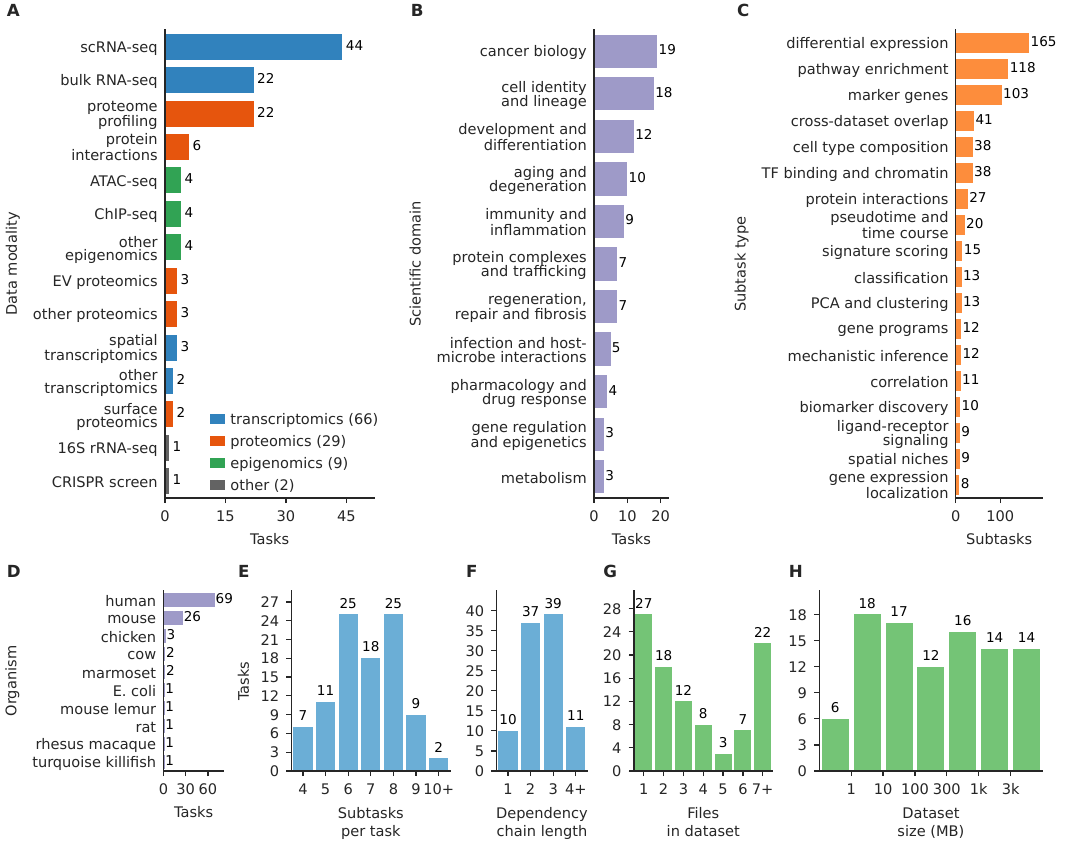}
\caption{\textbf{OmicsBench task statistics.} (\textbf{A}) Data modality, with a task counted once per modality it includes. (\textbf{B}) Scientific domain, one per task. (\textbf{C}) Subtask type, one per subtask. (\textbf{D}) Organism, with a task counted once per organism. (\textbf{E}) Number of subtasks per task. (\textbf{F}) Length of the longest dependency chain in each task's DAG. (\textbf{G}) Number of files in each task's dataset. (\textbf{H}) Total size of each task's dataset in MB, in log-spaced bins.}
\label{fig:composition}
\end{figure}

\section{Experiments}
\label{sec:experiments}

\subsection{Setup}
\label{sec:setup}

For this evaluation, we set the gate threshold $\tau$ to 0.7, the attempt limit $\theta$ to 3, the time limit per turn $t_{\max}$ to 1800~s, and the number of runs per task in each mode $k$ to 3 (\tabref{tab:parameters}).
We evaluated 12 models on three harnesses, with each harness pinned to one version and with reasoning effort set to high wherever the model allowed it (\appref{app:model-configurations}, \tabref{tab:models}).
All 12 models ran on OpenCode, and Opus 5 and GPT-5.6 Sol also ran on their native harnesses, Claude Code and Codex, for 14 model/harness combinations.
Of the three pairs of harnesses, two share a single model each and the third shares none, so we cannot make any claims as to the effect of different harnesses on performance.
Every run took place in its own Docker container with no network access (\appref{app:runtime-environment}).

We treat the 97 tasks as a sample from a larger population of bioinformatics tasks, so the uncertainty we report is how much a mean could move had a different 97 been drawn.
Because runs of the same task are not independent draws from that population, we averaged each task's three runs into one score before computing any statistic, as counting them separately would understate the uncertainty \citep{miller_adding_2024}.
Every 95\% confidence interval is the mean plus or minus 1.96 standard errors, computed from the sample standard deviation of the 97 task scores.

\subsection{Evaluation results}
\label{sec:main-results}

No model/harness combination comes close to saturating OmicsBench in either mode, with the best, Opus 5 on OpenCode, averaging 0.57 in single-pass mode and 0.67 in gated mode, and the weakest, Grok 4.3, averaging 0.43 and 0.54 (\figref[A]{fig:leaderboard}).
On their native harnesses, Opus 5 and GPT-5.6 Sol each score within 0.02 of their OpenCode results.
Gated retries raise every combination's score, by 0.09 to 0.13.
Within each combination, task scores span nearly the full range (\figref[A]{fig:supp-question-profiles}), and the three runs of a task vary little from one another (\figref{fig:supp-run-consistency}).
Task difficulty is spread rather than bunched, running from 0.24 to 0.94 in single-pass mode (\figref[B]{fig:leaderboard}), and gated retries lift the whole distribution without flattening it (\figref[C]{fig:leaderboard}, \figref{fig:supp-question-profiles}).
Agents drew on 103 distinct domain packages spanning Python, R, and command-line ecosystems, with a median of 2 to 9 per task depending on the combination (\figref{fig:supp-domain-packages}, \figref{fig:supp-package-dotplot}).

\begin{figure}[H]
\centering
\includegraphics[width=\textwidth]{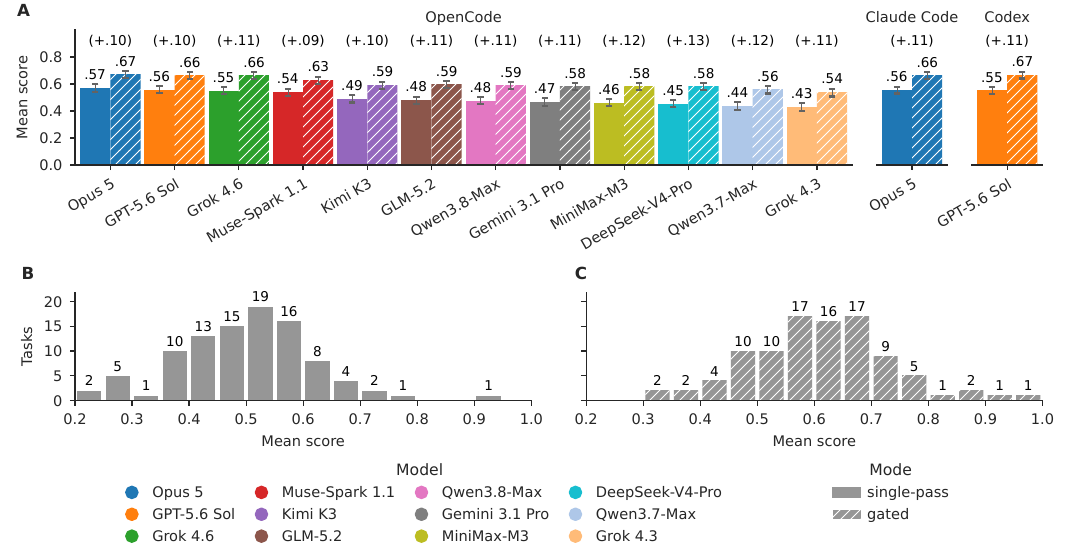}
\caption{\textbf{Performance of the 12 models and three harnesses.} (\textbf{A}) Mean score per combination in each mode, with the gated gain in parentheses above each pair. Error bars are 95\% confidence intervals over the 97 tasks. (\textbf{B, C}) Task difficulty, a task's mean score over all combinations, in each mode.}
\label{fig:leaderboard}
\end{figure}

\subsection{Cost and efficiency}
\label{sec:efficiency}

\begin{figure}[H]
\centering
\includegraphics[width=\textwidth]{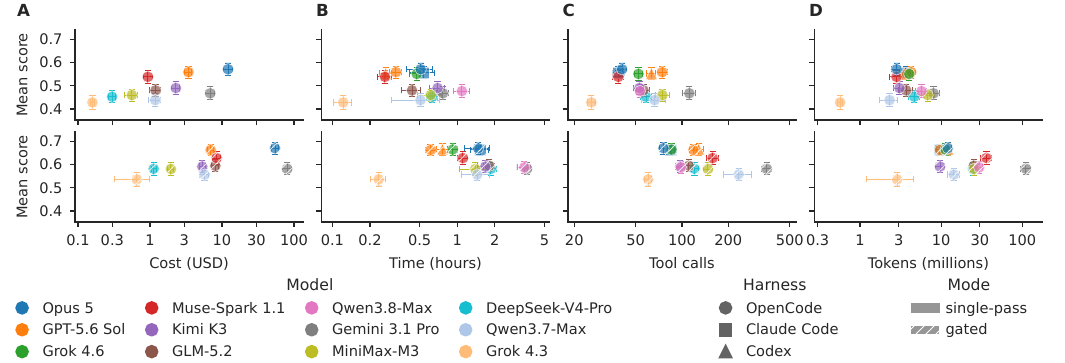}
\caption{\textbf{Cost and efficiency.} (\textbf{A}) Cost, (\textbf{B}) wall-clock time, (\textbf{C}) tool calls, and (\textbf{D}) tokens per task against mean score, in single-pass mode (top) and gated mode (bottom). Cost is available only on OpenCode, and not for Grok 4.6 or Qwen3.8-Max. Error bars are 95\% confidence intervals of mean score.}
\label{fig:efficiency}
\end{figure}

The model/harness combinations that we tested separate far more by efficiency than by performance (\figref{fig:efficiency}).
In single-pass mode, Gemini 3.1 Pro costs 12 times what MiniMax-M3 does while scoring only 0.01 higher, and 7 times what Muse-Spark 1.1 does while actually scoring 0.07 lower (\figref[A]{fig:efficiency}).
Effort has a similar relationship with performance, with Gemini 3.1 Pro making more tool calls and using more tokens\footnote{Token counts may not be directly comparable between harnesses, since each harness records token use in its own way.} than any other combination in both modes while still scoring in the bottom half of the combinations on OpenCode (\figref[C]{fig:efficiency}, \figref[D]{fig:efficiency}, \figref{fig:supp-token-composition}).
To the extent that OmicsBench represents the work it models, choosing an agent for it is as much an economic question as a capability one.
For each measure (cost, time, tool calls, or tokens), a Pareto-optimal model is defined as one that no other model on OpenCode outscores at an equal or lower value of that measure.
By this definition, Opus 5 and Grok 4.3 are Pareto-optimal on all four measures in both modes, as is Muse-Spark 1.1 in single-pass mode, while Gemini 3.1 Pro, GLM-5.2, and Qwen3.8-Max are optimal on none of the measures for which they have data (\figref{fig:efficiency}, \tabref{tab:pareto}).

\subsection{Performance across turns, dependencies, and retries}
\label{sec:turns-tries}

\begin{figure}[!b]
\centering
\includegraphics[width=\textwidth]{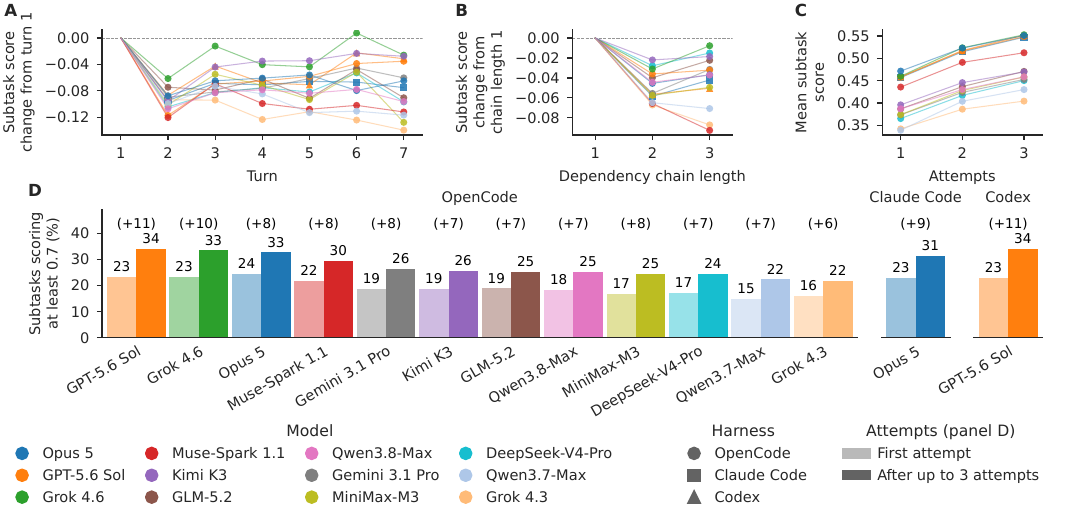}
\caption{\textbf{Subtask scores across a session and across gated retries.} (\textbf{A}) Change in mean subtask score from the first turn, in single-pass mode, through turn 7. (\textbf{B}) The same by dependency chain length, through length 3, as too few subtasks have longer chains for a stable average. (\textbf{C}) Mean best score within 1 to 3 gated attempts. (\textbf{D}) Share of subtasks reaching the gate ($\tau = 0.7$) on the first attempt and within three attempts.}
\label{fig:turns-tries}
\end{figure}

An agent's score on a subtask depends on where the subtask falls in its session.
Every combination loses 0.06 to 0.12 by the second turn, and later turns stay below the first in all but one case (\figref[A]{fig:turns-tries}).
Subtasks that build on an agent's earlier answers also score lower than those with no upstream dependence, for every combination (\figref[B]{fig:turns-tries}).
Gated retries recover part of that loss, as every combination's best score so far rises with each attempt, most from the first attempt to the second (\figref[C]{fig:turns-tries}).
Even so, only 22 to 34\% of subtasks reach the gate within three attempts, up from 15 to 24\% on the first attempt (\figref[D]{fig:turns-tries}).
These analyses leave out the 177 of 662 subtasks graded only by a closed-choice grader (\tabref{tab:grader-catalog}), because we anecdotally saw that agents tended to exhaust their options by guessing, which we regard as a sign of reward hacking.
\subsection{Failure modes}
\label{sec:failure-modes}

\begin{figure}[!b]
\centering
\includegraphics[width=\textwidth]{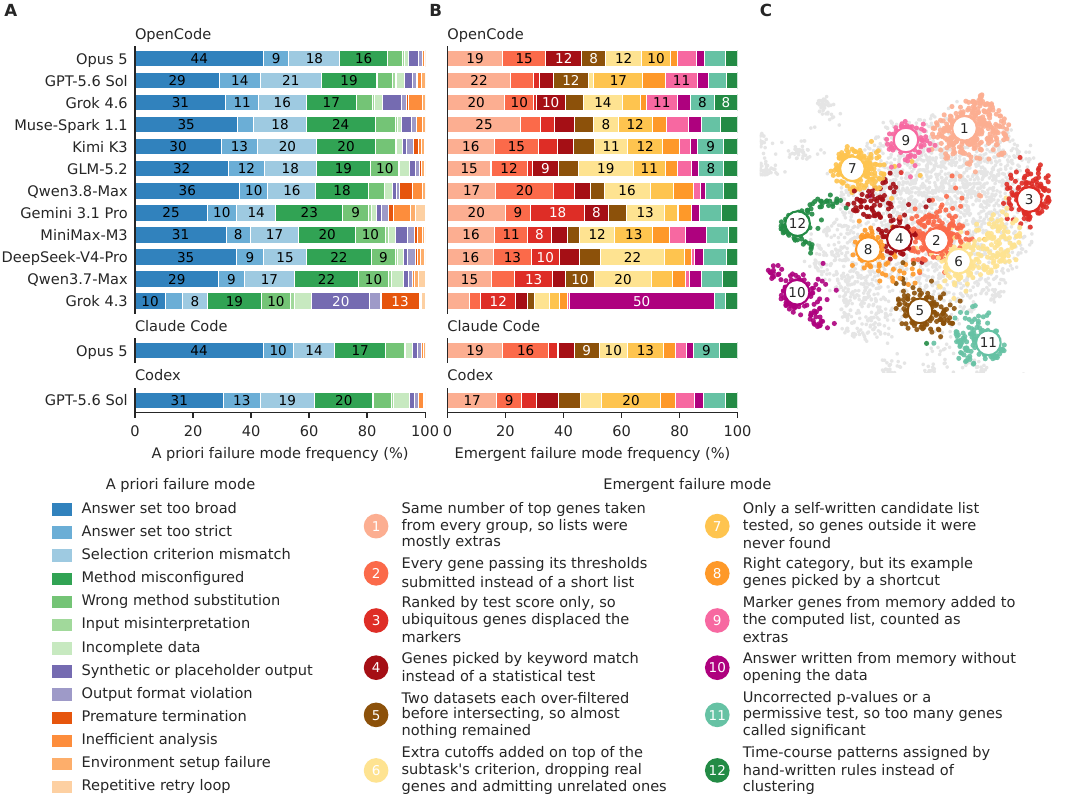}
\caption{\textbf{Failure modes, by model/harness combination.} (\textbf{A}) Each a priori mode's share of a combination's assigned modes. (\textbf{B}) Each emergent mode's share of a combination's clustered attempts. (\textbf{C}) UMAP of failure descriptions, one dot per attempt, colored by mode, gray if unassigned.}
\label{fig:failure-modes}
\end{figure}

The a priori failure mode distribution shows that, across all model/harness combinations, agents mostly fail in how they carry out the analysis rather than by failing to deliver an answer (\tabref{tab:apriori-modes}, \figref[A]{fig:failure-modes}).
Set-selection failures, in which the agent submits far more or far fewer results than the subtask requires, are the most common in every combination, at 25 to 71\% of assigned failure modes.
Grok 4.3 alone departs from this pattern, with synthetic or placeholder output at 20\% and writing no answer at 13\% of its assigned failure modes, against at most 7\% and 5\% for any other combination.
Where other agents filter their results poorly, Grok 4.3 more often fabricates them or gives up.

The 12 emergent failure modes cover 42\% of failed attempts and fall into five families of related mechanisms (\figref[C]{fig:failure-modes}, \secref{sec:failure-mode-analysis}, \tabref{tab:emergent-modes}), and most of the remaining 58\% are one-off mechanisms too rare to form a cluster, with the rest in the three clusters we set aside (\appref{app:emergent-modes}).
The families have similar shares across combinations (\figref[B]{fig:failure-modes}).
The largest, at 37 to 55\% of a combination's clustered attempts, is submitting more than the subtask asked for (\modedot{fcae91}{1}--\modedot{a50f15}{4}), for example by returning a fixed number of top genes for every group (\modedot{fcae91}{1}), and the next, at 22 to 34\%, is substituting the agent's own criterion for the subtask's (\modedot{fee391}{6}--\modedot{fe9929}{8}), for example by testing only a list of candidate genes the agent wrote itself rather than the whole dataset (\modedot{fec44f}{7}).
The other three are substituting a different procedure (\modedot{66c2a4}{11}--\modedot{238b45}{12}, 10 to 16\%), putting prior knowledge in place of the data (\modedot{f768a1}{9}--\modedot{ae017e}{10}, 3 to 15\%), and submitting less than the subtask requested (\modedot{8c510a}{5}, 4 to 12\%).
Grok 4.3 again departs from this pattern and is left out of these ranges, as writing the answer from memory (\modedot{ae017e}{10}) alone is half of its clustered attempts, against at most 7\% for any other combination.
\section{Discussion}
\label{sec:discussion}

\paragraph{Implications for training.} OmicsBench points to three behaviors worth targeting in training.
First, agents need to base their answers on the criteria that tasks state rather than on their own interpretation of them (\secref{sec:failure-modes}).
Second, they need to stay grounded in the scientific context across a session, since every model/harness combination scores lower the further a subtask sits from the start of the session and from the original data, both after its first turn and on subtasks that build on its earlier answers (\secref{sec:turns-tries}).
Third, they need to make better use of in-context feedback, as three attempts still leave most subtasks short of the gate, with the first retry worth more than the second (\secref{sec:turns-tries}).

\paragraph{Limitations.} The most important limitation is that gated mode is open to reward hacking by design.
Any signal that an answer has failed lets an agent search for an answer that passes rather than correct the analysis behind it, and this holds for every kind of feedback on every kind of subtask.
We saw it directly on closed-choice subtasks, where agents tended to exhaust the options by guessing, and so left those subtasks out of the session and retry analyses (\secref{sec:turns-tries}), but subtler forms are possible on any subtask, such as loosening a threshold until a gene list passes or resubmitting small variants of a failing answer.
The gated gains are therefore an upper bound on how much in-context feedback helps agents recover, and the share of those gains owed to reward hacking is unquantified in this study.
Furthermore, with 97 tasks across 11 biological domains and 18 subtask analysis types, each domain or analysis type holds too few tasks to compare models within it (\figref[B]{fig:composition}, \figref[C]{fig:composition}).
Only Opus 5 and GPT-5.6 Sol ran on more than one harness, so the effect of the harness cannot be separated from that of the model.
Lastly, the LLM judge was tested for reproducibility but not against human annotators.

\paragraph{Future work.} The clearest next step for long-horizon, multi-turn evaluation is a simulated user that answers an agent's questions and revises its requests as a scientist would.
Such a user, however, would have to be held to accurate replies, and it would cost the evaluation its determinism.
Equally as pressing is measuring and limiting reward hacking under feedback, for example by comparing gated gains against retries that receive no feedback, by auditing retry transcripts for agents searching the answer space, and by designing feedback that points to the faulty step without revealing what a passing answer looks like.
Additionally, more tasks in each domain and analysis type would allow models to be compared within them, and human annotation of failed attempts would test the LLM judge against human judgment.

\section{Conclusion}
\label{sec:conclusion}

We introduced OmicsBench, 97 research-study-scale bioinformatics tasks whose 662 subtasks are posed to agents as the turns of one session per task, which lets it separate a subtask's position in the session from its place in a dependency chain in a way that single-prompt benchmarks cannot.
We release the runner and graders under the MIT license and three tasks under CC BY 4.0.
Across 12 models on three harnesses, no combination comes close to saturating the benchmark, every combination scores lower after its first turn and on subtasks that build on its earlier answers, and gated retries recover only part of that loss.
Both of our failure analyses point to the same shortcoming in how agents carry out bioinformatics analyses, namely that they follow their own habits rather than the subtask's instructions, most often by choosing an answer set with a criterion of their own rather than the one the subtask states.

\raggedbottom
\bibliographystyle{afterquery}
\bibliography{references}

\appendix
\appendixcontents
\makeatletter
\@addtoreset{table}{section}
\@addtoreset{figure}{section}
\makeatother
\renewcommand{\thetable}{\thesection\arabic{table}}
\renewcommand{\thefigure}{\thesection\arabic{figure}}
\renewcommand{\theHtable}{\thesection.\arabic{table}}
\renewcommand{\theHfigure}{\thesection.\arabic{figure}}

\section{Supplementary figures and tables}
\label{app:supp-figures}

\begin{figure}[H]
\centering
\includegraphics[width=\textwidth]{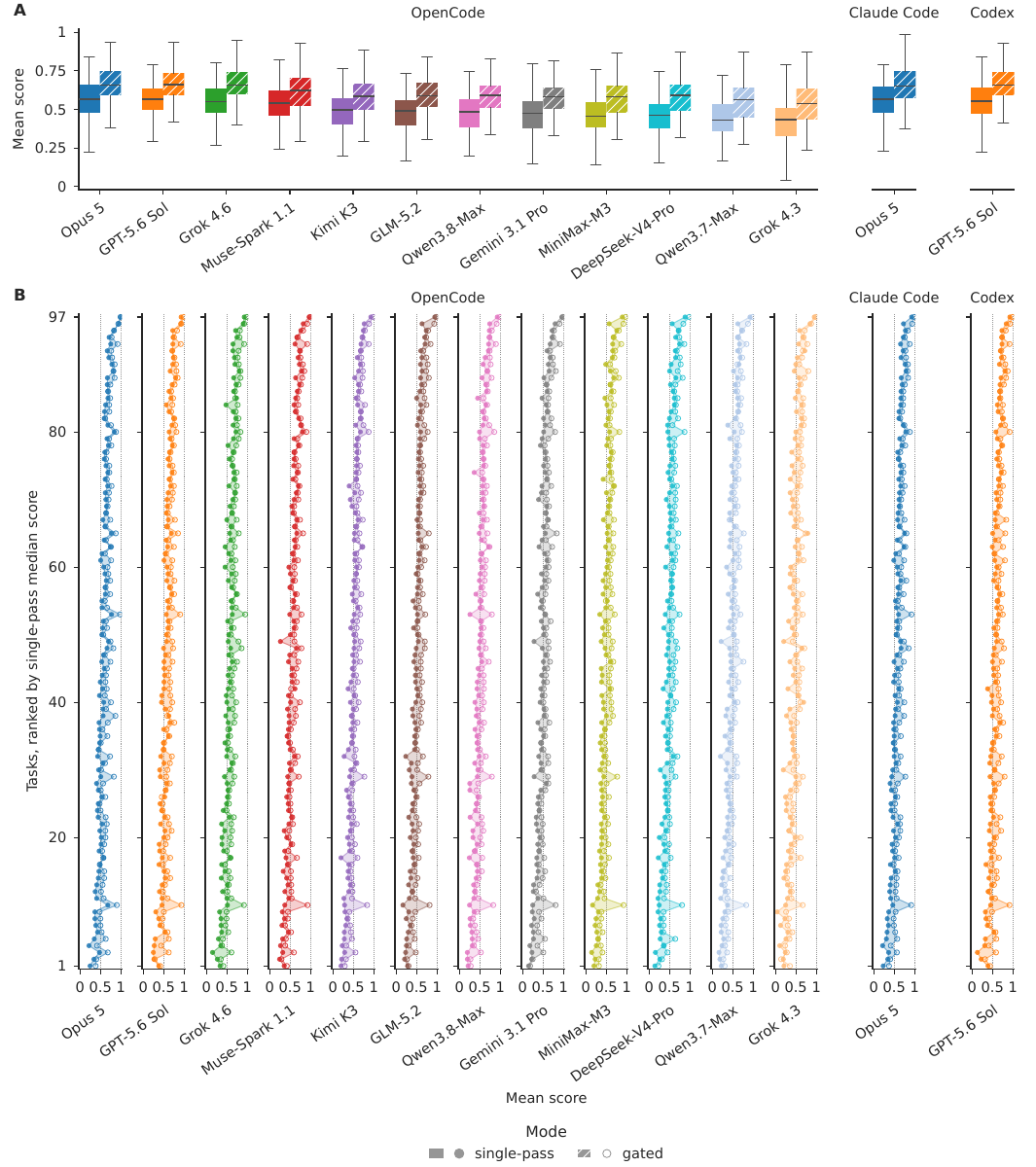}
\caption{\textbf{Per-task scores by model/harness combination.} (\textbf{A}) Distribution of the 97 task scores per combination in each mode. (\textbf{B}) One row per task, ranked by median single-pass score across combinations, hardest at the bottom. The colored band between dots is tinted where gating raises that task's score for that combination and left white where it lowers it.}
\label{fig:supp-question-profiles}
\end{figure}

\begin{figure}[H]
\centering
\includegraphics[width=\textwidth]{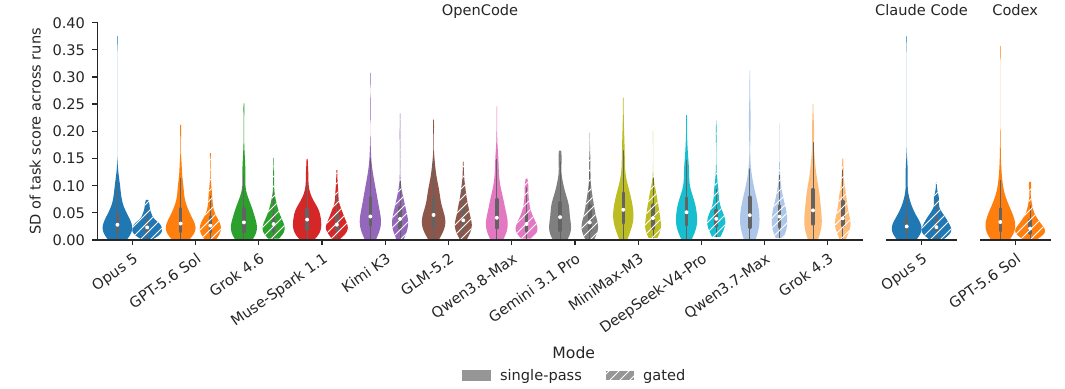}
\caption{\textbf{Run-to-run consistency.} Each violin covers one model/harness combination in one mode and shows, over the 97 tasks, the distribution of the standard deviation of a task's score across its three runs. The inner bar marks the interquartile range and the white dot marks the median.}
\label{fig:supp-run-consistency}
\end{figure}

\begin{figure}[H]
\centering
\includegraphics[width=\textwidth]{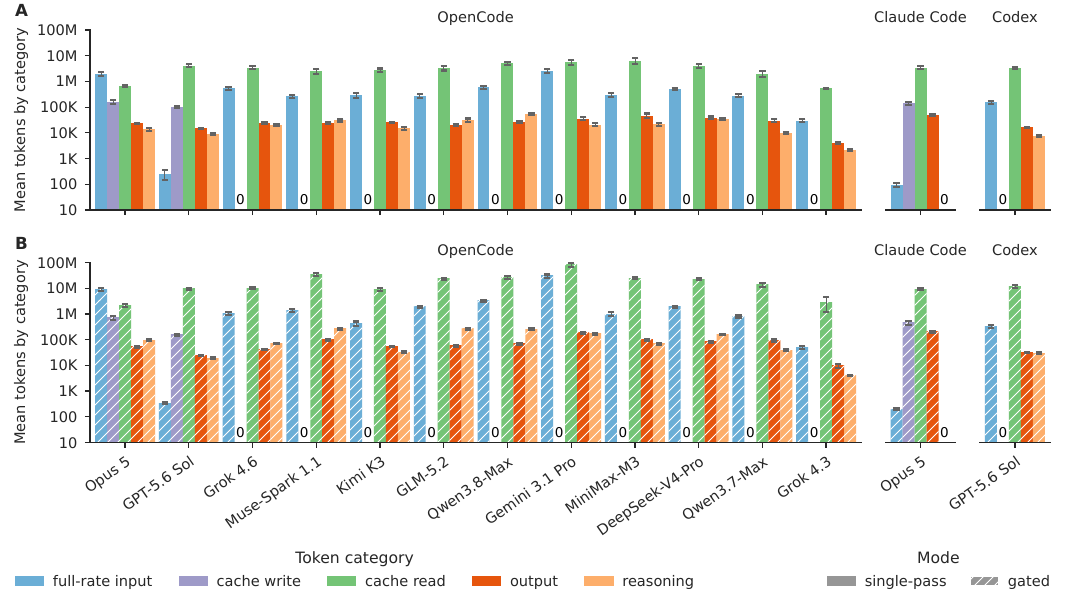}
\caption{\textbf{Tokens per task by category.} (\textbf{A}) Single-pass mode. (\textbf{B}) Gated mode. Each bar is a combination's mean tokens per task in one category, on a logarithmic axis. Error bars are 95\% confidence intervals.}
\label{fig:supp-token-composition}
\end{figure}

\begin{figure}[H]
\centering
\includegraphics[width=\textwidth]{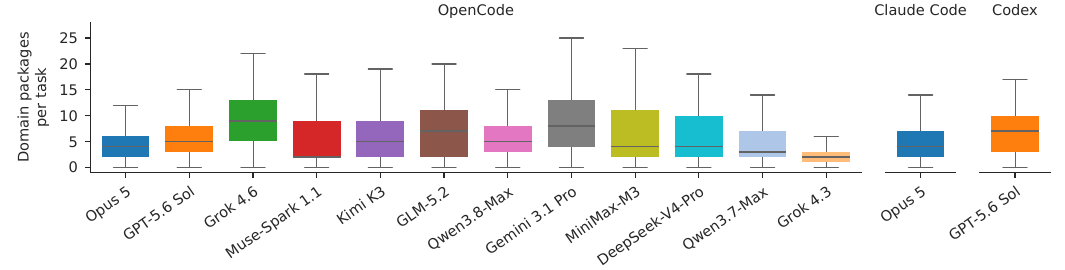}
\caption{\textbf{Domain packages per task.} Each box covers one model/harness combination and shows, over the 97 tasks, the number of distinct domain packages the combination invoked on a task across its six runs, three in each mode.}
\label{fig:supp-domain-packages}
\end{figure}

\begin{figure}[H]
\centering
\includegraphics[width=\textwidth]{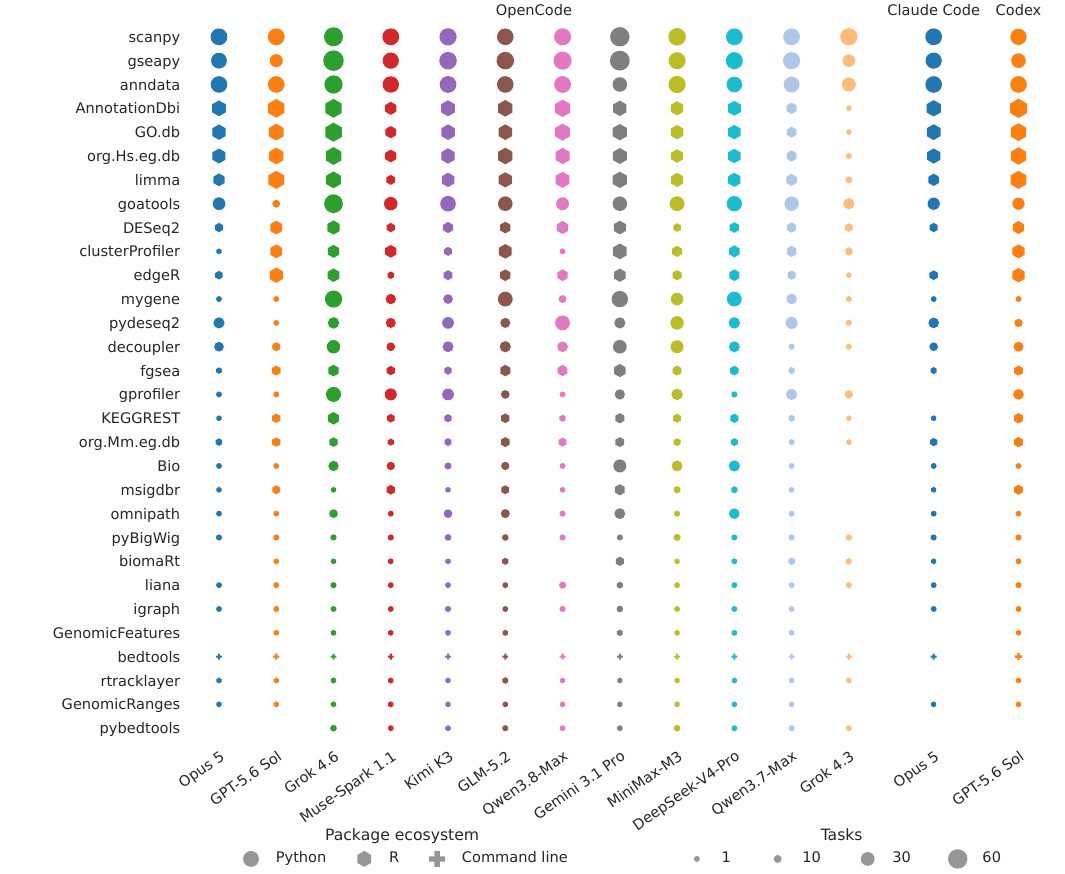}
\caption{\textbf{Domain packages invoked by each model/harness combination.} Each row is one of the 30 most used domain packages, ordered by the number of tasks on which it was invoked, summed over all combinations (most used at the top). Marker area is the number of the 97 tasks on which that combination invoked the package across its six runs.}
\label{fig:supp-package-dotplot}
\end{figure}

\begin{table}[H]
\centering
\caption{\textbf{Pareto-optimality of models on OpenCode.} A check marks a model that is Pareto-optimal on a measure in one mode, meaning that no other model on OpenCode, the one harness on which all 12 models ran, scores higher at the same or lower value of that measure. N/A marks missing data.}
\label{tab:pareto}
\footnotesize
\begin{tblr}{
  width = \textwidth,
  colspec = {Q[l] *{8}{X[c]}},
  colsep = 3pt,
  rows = {valign=m},
  hline{1,Z} = {0.08em},
  hline{3} = {0.05em},
  hline{2} = {2-3,4-5,6-7,8-9}{leftpos=-0.6, rightpos=-0.6, 0.03em},
  rowsep = 2pt,
}
\SetCell[r=2]{l} Model & \SetCell[c=2]{c} Cost & & \SetCell[c=2]{c} Time & & \SetCell[c=2]{c} Tool calls & & \SetCell[c=2]{c} Tokens & \\
 & Single-pass & Gated & Single-pass & Gated & Single-pass & Gated & Single-pass & Gated \\
Opus 5 & $\checkmark$ & $\checkmark$ & $\checkmark$ & $\checkmark$ & $\checkmark$ & $\checkmark$ & $\checkmark$ & $\checkmark$ \\
GPT-5.6 Sol & $\checkmark$ & $\checkmark$ & $\checkmark$ & $\checkmark$ &  &  &  & $\checkmark$ \\
Gemini 3.1 Pro &  &  &  &  &  &  &  &  \\
GLM-5.2 &  &  &  &  &  &  &  &  \\
Grok 4.3 & $\checkmark$ & $\checkmark$ & $\checkmark$ & $\checkmark$ & $\checkmark$ & $\checkmark$ & $\checkmark$ & $\checkmark$ \\
Grok 4.6 & N/A & N/A &  & $\checkmark$ &  &  &  & $\checkmark$ \\
Kimi K3 &  & $\checkmark$ &  &  &  &  &  &  \\
MiniMax-M3 & $\checkmark$ &  &  &  &  &  &  &  \\
Muse-Spark 1.1 & $\checkmark$ &  & $\checkmark$ &  & $\checkmark$ &  & $\checkmark$ &  \\
Qwen3.7-Max &  &  &  &  &  &  & $\checkmark$ &  \\
Qwen3.8-Max & N/A & N/A &  &  &  &  &  &  \\
DeepSeek-V4-Pro & $\checkmark$ & $\checkmark$ &  &  &  &  &  &  \\
\end{tblr}
\end{table}

\section{Notation and parameters}
\label{app:notation}

We set $\tau = 0.7$ as a compromise between a threshold of 1, which would fail answers with a single extra or missing member, and a threshold at or below 0.5, which would pass answers that are mostly wrong.
Setting $\theta = 3$ bounds the cost of gated mode at 3 times that of single-pass mode and mirrors the review-and-revise cycle a scientist would tolerate before taking over.
The time limit $t_{\max} = 1800$~s exceeds the typical turn time by a wide margin, so a timeout reflects unproductive work rather than a tight budget.
Three runs per task ($k = 3$) is the smallest replicate design that lets us report run-to-run variance (\figref{fig:supp-run-consistency}) while keeping the total at 8,148 sessions (97 tasks, 14 model/harness combinations, two modes, and three runs).

\begin{table}[H]
\centering
\caption{\textbf{Notation.} Symbols used in the OmicsBench formulation.}
\label{tab:notation}
\small
\begin{tblr}{
  width = \textwidth,
  colspec = {Q[l] X[l]},
  rows = {valign=m},
  hline{1,Z} = {0.08em},
  hline{2} = {0.05em},
  rowsep = 2pt,
}
Symbol & Meaning \\
$T = (P, D, G)$ & A task, composed of a task prompt $P$, biological dataset $D$, and DAG $G$ \\
$P$ & The task prompt of a task \\
$D$ & The biological dataset of a task \\
$G = (V, E)$ & The task DAG, with nodes $V$ and edges $E$ \\
$v \in V$ & A node of the task DAG, one subtask \\
$e = (u, v) \in E$ & An edge of the task DAG, meaning that subtask $v$ depends on the answer to subtask $u$ \\
$v = (v_p, v_e)$ & A subtask: its prompt $v_p$ and its evaluation set $v_e$ \\
$v_p$ & Subtask prompt \\
$v_e=\{(g_1, y_1), (g_2, y_2), \ldots\}$ & The graders of subtask $v$, each paired with its answer key \\
$a_v$ & The agent's answer file for subtask $v$ \\
$g_i(a_v, y_i) \in [0, 1]$ & The score of grader $g_i$ on its field in $a_v$ \\
$s_v \in [0, 1]$ & The score of subtask $v$, the mean of its grader scores, $\frac{1}{|v_e|} \sum_i g_i(a_v, y_i)$ \\
$S$ & The score of a task, the mean of its subtask scores, $\frac{1}{|V|} \sum_{v \in V} s_v$ \\
\end{tblr}
\end{table}

\begin{table}[H]
\centering
\caption{\textbf{Benchmark parameters.} OmicsBench settings and their values in this study.}
\label{tab:parameters}
\small
\begin{tblr}{
  width = \textwidth,
  colspec = {X[l] Q[l] Q[l]},
  rows = {valign=m},
  hline{1,Z} = {0.08em},
  hline{2} = {0.05em},
  rowsep = 2pt,
}
Parameter & Symbol & Value \\
Gate threshold in gated mode & $\tau$ & 0.7 \\
Attempt limit per subtask in gated mode & $\theta$ & 3 \\
Time limit per turn (s) & $t_{\max}$ & 1800 \\
Runs per task in each mode & $k$ & 3 \\
\end{tblr}
\end{table}

\section{Benchmark prompts}
\label{app:benchmark-prompts}

\subsection{Benchmark system prompt and task prompt}
\label{app:system-prompt}

\begin{promptverb}
You are a computational biologist. You have access to a computational
environment with standard bioinformatics tools.

Your workspace is organized as follows:
- /data/                -- input data files (read-only)
- /workspace/answers/   -- write ONLY your graded answer files here (JSON)
- /workspace/scratch/   -- working area for intermediate files, scripts, etc.

You will receive a series of analysis subtasks. For each subtask, you will
be told what to do and where to write your answer. Complete each subtask
before moving on to the next.

How to succeed:

- Work autonomously. Do not ask for clarification. If an instruction is
  ambiguous, note your interpretation in your reasoning, pick the most
  reasonable one, and proceed.

- The graded artifact is the file you write at
  /workspace/answers/<requested_filename>, not anything you print in your
  response. If you describe the answer but do not write it to the correct
  file, you score zero. Writing the file is the job.

- Write exactly to the path and filename the subtask specifies. Do not
  rename, pluralize, change the extension, or route the file elsewhere. If a
  subtask asks for foo.json, the answer lives at /workspace/answers/foo.json.

- Answer files are strict JSON. No comments, no trailing commas, no markdown
  fences, no wrapping the answer in extra keys. The file's top-level
  structure must match the schema in the subtask literally.

- For multiple-choice answers, write the selected answer exactly as one of
  the choices shown in the subtask. Letter case is ignored by the grader, but
  spelling, punctuation, and qualifiers must match the listed choice.

- Before writing code that processes an input file, inspect it (shape,
  columns, dtypes, first few rows). Do not assume structure from memory.

- Commit to your best answer. Do not write placeholder values like "TBD",
  "uncertain", or null when the subtask asks for a concrete value.

- Do not delete files in /workspace/scratch/ between subtasks; later
  subtasks may depend on them.

Below, you will find a description of the data that you will be using to accomplish your subtasks. Read through it carefully before starting any analysis.

<task prompt>

Data modalities:
- <modality>

Organisms:
- <organism>

Files available in /data/:

1. <file name>
   - Description: <file description>
   - Format: <format>
   - Data modalities: <modality>
   - Organisms: <organism>
   - Platforms: <platform>
   - Path: /data/<file name>.<format>
\end{promptverb}

\subsection{Subtask prompt}
\label{app:prompt-task}

\begin{promptverb}
Subtask: <subtask name>

<subtask instruction>

Write your answer as JSON to /workspace/answers/<subtask name>.json. The schema below uses placeholders like ``<integer>``, ``<string>``, ``<number>``, ``<boolean>``, and ``<any>``. You MUST replace every placeholder with an actual value of that type. Do NOT output the literal placeholder string (``<integer>`` as a string value is never correct). The shape around the placeholder(s) (brackets, braces, keys, and list wrappers) is literal.

<answer schema>
\end{promptverb}

In gated mode, a subtask whose predecessor took more than one attempt carries a note after the subtask line, naming the workspace it inherits:

\begin{promptverb}
Note: Your workspace has been restored to the state from attempt <n> of the previous subtask (<previous subtask name>), which was your best attempt. Files on disk may differ from your most recent attempt.
\end{promptverb}

When the runner includes the numeric score in its feedback (\appref{app:prompt-retry}), the note reads instead:

\begin{promptverb}
Note: Your workspace has been restored to the state from attempt <n> of the previous subtask (<previous subtask name>), which had the best score (<score>). Files on disk may differ from your most recent attempt.
\end{promptverb}

When every attempt at the predecessor failed, the note reads instead:

\begin{promptverb}
Note: Every attempt at the previous subtask (<previous subtask name>) failed. The workspace may be in an inconsistent state; verify file contents before relying on them.
\end{promptverb}

A subtask whose answer carries group labels appends this block:

\begin{promptverb}
Grouped fields:
Use the listed group labels exactly for each grouped answer field.

- "<field name>": ["<group label>", ...]
\end{promptverb}

A subtask whose answer carries a closed set of choices appends this one instead:

\begin{promptverb}
Choice fields:
Multiple-choice fields are scalar values. Multi-select fields are lists. For list-valued fields, include every correct choice and no unsupported choices. Answer exactly as one of the listed choices. Letter case is ignored by the grader for string choices, but spelling, punctuation, and qualifiers must match the listed choice.

- "<field name>": ["<choice>", ...]
\end{promptverb}

A subtask whose answer carries a labeled numeric matrix appends this one:

\begin{promptverb}
Labeled matrix fields:
Use these row and column labels exactly for each labeled numeric matrix field.

- "<field name>" rows: ["<row label>", ...]
- "<field name>" columns: ["<column label>", ...]
\end{promptverb}

\subsection{Retry prompt}
\label{app:prompt-retry}

Below is the default retry prompt.
\begin{promptverb}
Your previous attempt scored below the required threshold. Please review
your approach and try again. Write your updated answer to
/workspace/answers/<subtask name>.json.
\end{promptverb}

An attempt that cannot be graded at all is returned with the reason instead:

\begin{promptverb}
Your previous attempt could not be scored. Reason: <reason>

Please fix the issue and write a valid answer to
/workspace/answers/<subtask name>.json.
\end{promptverb}

The runner can also include the numeric score in its feedback.
In that form, the retry prompt reads as follows, the unscorable form is unchanged, and the workspace-restore note in \appref{app:prompt-task} gives the kept attempt's score.

\begin{promptverb}
Your previous attempt scored <score> out of 1.0, which is below the
required threshold. Please review your approach and try again. Write your
updated answer to /workspace/answers/<subtask name>.json.
\end{promptverb}

\section{Example task}
\label{app:example-question}

\paragraph{Example task prompt}

\begin{quote}
The adult human kidney cortex was profiled with paired single-nucleus RNA sequencing (transcriptome) and single-nucleus ATAC sequencing (chromatin accessibility, provided as a per-cell gene-activity matrix plus raw fragments). Both single-nucleus datasets are annotated with cell-type and cell-state labels, including a rare VCAM1-expressing proximal tubule subpopulation (labeled PT\_VCAM1) that sits alongside the remaining, differentiated proximal tubule. The scientific goal is to reconstruct the molecular identity and regulation of this VCAM1+ proximal tubule state from the data: define its transcriptional signature relative to the remaining proximal tubule, determine how the genes that drive its emergence change and whether those changes are mirrored at the chromatin-accessibility level, determine which transcription-factor programs switch as the state appears, and characterize the biological programs it activates. Use the annotated proximal tubule cells in the provided matrices to derive these results.
\end{quote}

\begin{table}[H]
\centering
\caption{\textbf{Files in the example task's dataset.}}
\label{tab:example-data}
\small
\begin{tabular}{@{}p{4.9cm} p{\dimexpr\textwidth-4.9cm-2\tabcolsep\relax}@{}}
\toprule
File & Description \\
\midrule
\texttt{snrna\_expression.h5ad} & Single-nucleus RNA-seq of adult human kidney cortex: log-normalized expression matrix (cells $\times$ genes) with per-cell cell-type/cell-state annotations in \texttt{obs} (\texttt{author\_cell\_type} includes PT for the remaining proximal tubule and PT\_VCAM1 for the VCAM1+ proximal tubule subpopulation, plus other nephron cell types). Gene symbols are in \texttt{var['feature\_name']}. \\
\addlinespace[2pt]
\texttt{snatac\_gene\_activity.h5ad} & Single-nucleus ATAC-seq of adult human kidney cortex represented as a per-cell gene-activity matrix (log-scaled accessibility aggregated over each gene's body and promoter; cells $\times$ genes) with per-cell cell-type/cell-state annotations in \texttt{obs} (\texttt{author\_cell\_type} includes PCT and PST for proximal tubule segments and PT\_VCAM1 for the VCAM1+ proximal tubule subpopulation, plus other nephron cell types). Gene symbols are in \texttt{var['feature\_name']}. \\
\addlinespace[2pt]
\texttt{snatac\_fragments.tsv.gz} & Raw single-nucleus ATAC-seq fragments for the same kidney cortex nuclei, block-compressed TSV keyed by genomic coordinate and cell barcode. Needed for peak- or motif-level analyses such as transcription-factor motif activity; the gene-activity matrix suffices for gene-level accessibility comparisons. \\
\addlinespace[2pt]
\texttt{snatac\_\allowbreak fragments.tsv.gz.tbi} & Tabix index for the single-nucleus ATAC-seq fragments file, enabling coordinate-range queries. \\
\bottomrule
\end{tabular}
\end{table}

\begin{longtable}{@{}>{\small}p{0.3cm} >{\fontsize{8}{9.6}\selectfont\raggedright\arraybackslash}p{2.4cm} >{\fontsize{8}{9.6}\selectfont\raggedright\arraybackslash}p{2.4cm} >{\small}p{\dimexpr\textwidth-5.1cm-6\tabcolsep\relax}@{}}
\caption{\textbf{Subtask prompts of the example task.} Numbers on the left refer to the boxes in the flowchart of \figref[A]{fig:protocol}, which omits \texttt{pt\_vcam1\_programs}.}\label{tab:example-tasks} \\
\toprule
 & \small Subtask & \small Depends on & Subtask prompt \\
\midrule
\endfirsthead
 & \small Subtask & \small Depends on & Subtask prompt \\
\midrule
\endhead
\bottomrule
\endlastfoot
\modedot{636363}{1} & \texttt{pt\_\allowbreak vcam1\_\allowbreak signature} & Root subtask, no upstream subtasks & Using the single-nucleus RNA-seq expression matrix, contrast the VCAM1+ proximal tubule subpopulation (PT\_VCAM1) against the remaining proximal tubule (PT) and derive the differential transcriptional signature of this cell state, testing genes detected in at least 20\% of cells in either group with a log fold-change threshold of 0.25 and Bonferroni-adjusted $p < 0.05$. Report the 15 most strongly up-regulated and, separately, the 15 most strongly down-regulated genes in PT\_VCAM1, ranked by average log fold-change as computed by Seurat's FindMarkers (avg\_logFC: the natural log of one plus the mean of expm1(expression) in PT\_VCAM1, minus the same quantity in PT, on the provided log-normalized values). Provide two lists of gene symbols keyed as `upregulated' and `downregulated'. \\
\addlinespace[3pt]
\modedot{636363}{2} & \texttt{pt\_\allowbreak vcam1\_\allowbreak transition\_\allowbreak drivers} & \texttt{pt\_\allowbreak vcam1\_\allowbreak signature} & Order the proximal tubule cells along a transition from the differentiated proximal tubule (PT) to the VCAM1+ state (PT\_VCAM1). The source study reports four hallmark genes whose expression changes define this transition: VCAM1, TPM1, SLC5A12, and SLC4A4. Determine the direction in which each of them changes along the transition and report them grouped by direction: genes induced (rising toward PT\_VCAM1) under group label `induced', and genes repressed (falling toward PT\_VCAM1) under group label `repressed'. Each group holds gene symbols. \\
\addlinespace[3pt]
\modedot{636363}{3} & \texttt{pt\_\allowbreak vcam1\_\allowbreak regulators} & \texttt{pt\_\allowbreak vcam1\_\allowbreak signature} & Nominate the transcription-factor programs that switch as the proximal tubule adopts the VCAM1+ state by computing transcription-factor motif activity from the single-nucleus ATAC fragments (for example chromVAR deviations over JASPAR vertebrate motifs), comparing PT\_VCAM1 nuclei with the differentiated proximal tubule nuclei (the PCT and PST segments). Identify the transcription factor whose motif activity is most clearly lost in PT\_VCAM1 and the two transcription factors whose motif activity is most strongly gained. You may corroborate the direction using the PT\_VCAM1 signature you derived, transcription-factor gene expression, and gene activity. Where motif activity alone cannot separate paralogs that share a near-identical motif, report the paralog whose own expression and gene activity move in the same direction as the motif activity. Report each regulator by its own gene symbol, grouped as `repressed\_regulators' (lost in PT\_VCAM1) and `induced\_regulators' (gained in PT\_VCAM1). \\
\addlinespace[3pt]
 & \texttt{pt\_\allowbreak vcam1\_\allowbreak programs} & \texttt{pt\_\allowbreak vcam1\_\allowbreak signature} & Run a functional-enrichment analysis on the full PT\_VCAM1 signature, up- and down-regulated genes together, and select the biological programs that are enriched in this subpopulation relative to the remaining proximal tubule. Choose all applicable programs from the provided options. \\
\addlinespace[3pt]
\modedot{636363}{4} & \texttt{pt\_\allowbreak vcam1\_\allowbreak accessibility\_\allowbreak concordance} & \texttt{pt\_\allowbreak vcam1\_\allowbreak transition\_\allowbreak drivers} & Using the single-nucleus ATAC gene-activity matrix, test whether the transition-driver genes identified for the PT-to-PT\_VCAM1 transition show concordant changes in chromatin accessibility. Compare gene activity in PT\_VCAM1 cells against the differentiated proximal tubule cells (the PCT and PST segments) and report which of those driver genes show increased accessibility in PT\_VCAM1 under group label `increased\_accessibility' and which show decreased accessibility under group label `decreased\_accessibility'. Each group holds gene symbols. \\
\end{longtable}

\section{Code and data availability}
\label{app:code-data-availability}

The OmicsBench runner and three of the 97 tasks with their datasets are available at \url{https://huggingface.co/datasets/gideonshaked/OmicsBench}, and the identifier databases against which the graders match at \url{https://huggingface.co/datasets/gideonshaked/OmicsBench-grader-databases}.
The runner, comprising the evaluation harness and every grader, is released under an MIT license, and the public tasks include the example task in \figref{fig:protocol} and \appref{app:example-question}.
The three public tasks, including their answer keys, are released under a CC BY 4.0 license, their datasets under the licenses of the studies that released them, and the identifier databases under the licenses of their source databases (\tabref{tab:identifier-databases}).
The remaining tasks are held back as a private test set to keep the benchmark useful for evaluating future models, and researchers who wish to evaluate an agent on these tasks may contact the authors.

\section{Grader catalog}
\label{app:grader-catalog}

\begin{longtable}{@{}>{\small}p{5.2cm} >{\small}p{\dimexpr\textwidth-5.2cm-2\tabcolsep\relax}@{}}
\caption{\textbf{OmicsBench grader catalog.}}\label{tab:grader-catalog} \\
\toprule
Grader & Description \\
\midrule
\endfirsthead
Grader & Description \\
\midrule
\endhead
\bottomrule
\endlastfoot
\texttt{absolute\_numeric\_tolerance} & Compares one numeric value using an absolute error band. \\
\texttt{abundance\_table} & Scores an unordered table of entity names paired with numeric abundances or quantities. \\
\texttt{amino\_acid\_set} & Scores an unordered set of amino acids using one-letter, three-letter, and full-name normalization. \\
\texttt{amr\_gene\_set} & Scores an unordered set of antimicrobial resistance genes using CARD ARO synonym resolution. \\
\texttt{case\_insensitive\_string} & Compares one scalar string after lowercasing both sides. \\
\texttt{case\_insensitive\_string\_list} & Scores an unordered set of literal strings while ignoring case. \\
\texttt{category\_count\_table} & Grades categorical metadata summaries such as cell-type counts, tissue compositions, sample-group sizes, or condition breakdowns. \\
\texttt{coordinate\_point\_set} & Scores unordered 2D or 3D point records by Euclidean distance, optionally requiring a point identifier to match before distance is scored. \\
\texttt{curve\_fit\_params} & Scores a four-parameter response curve fit using parameter-specific numeric tolerances. \\
\texttt{differential\_feature\_set} & Scores unordered differential-result rows keyed by feature with effect size and significance statistics. \\
\texttt{distribution} & Compares keyed numeric distributions after exact key alignment. \\
\texttt{docking\_pose\_set} & Scores unordered docking-pose records with ligand identity, docking score, and a nested set of interacting residues. \\
\texttt{enrichment\_result\_set} & Scores unordered enrichment-result rows keyed by term with significance and optional count statistics. \\
\texttt{entity\_pair\_set} & Scores unordered directed pair records with exact string endpoints. \\
\texttt{exact\_string} & Compares one scalar string with no normalization. \\
\texttt{exact\_string\_list} & Scores an unordered set of case-sensitive literal strings. \\
\texttt{gene\_pair\_set} & Scores unordered directed gene-gene pair records. \\
\texttt{gene\_set} & Scores an unordered set of gene names using NCBI Gene synonym resolution for the configured organism. \\
\texttt{gene\_taxon\_link\_set} & Scores unordered records linking an exact gene identifier to a taxon. \\
\texttt{grouped\_entity\_set} & Scores records where each group label owns an unordered set of entity or label items. \\
\texttt{grouped\_ranked\_list} & Scores records where each group label owns an ordered list of entity or label items. \\
\texttt{interval} & Scores one genomic interval by exact chromosome match and reciprocal span overlap. \\
\texttt{interval\_pair\_set} & Scores an unordered set of paired genomic loci. \\
\texttt{interval\_set} & Scores an unordered set of genomic intervals. \\
\texttt{labeled\_numeric\_matrix} & Scores a two-axis numeric table represented as unordered matrix cells. \\
\texttt{labeling} & Compares two parallel label assignments as relabel-invariant clusters. \\
\texttt{metabolite\_set} & Scores an unordered set of metabolite names using HMDB synonym resolution. \\
\texttt{mirna\_set} & Scores an unordered set of miRNA names using miRBase normalization. \\
\texttt{multiple\_choice} & Scores one scalar answer against a fixed homogeneous set of allowed choices. \\
\texttt{multi\_select\_choice} & Scores a list of selected choices against a fixed homogeneous set of allowed choices. \\
\texttt{numeric\_list} & Scores an ordered numeric vector either by per-position tolerance or by correlation. \\
\texttt{numeric\_list\_mean} & Scores a fixed-order numeric vector by averaging per-position absolute tolerance scores. \\
\texttt{numeric\_tolerance} & Compares one numeric value using absolute, relative, or log-scale tolerance. \\
\texttt{pair\_set} & Scores unordered records with two string/entity endpoints named ``a'' and ``b''. \\
\texttt{protein\_site\_set} & Scores unordered records that identify a protein/gene and residue site, optionally with a semantic site label. \\
\texttt{ranked\_amino\_acid\_list} & Scores an ordered amino-acid list after one-letter, three-letter, and full-name normalization. \\
\texttt{ranked\_amr\_list} & Scores an ordered list of AMR genes after CARD normalization. \\
\texttt{ranked\_gene\_list} & Scores an ordered gene list after gene synonym resolution. \\
\texttt{ranked\_list} & Scores an ordered list of strings after resolving each element with a selected string matcher. \\
\texttt{ranked\_metabolite\_list} & Scores an ordered list of metabolites after HMDB normalization. \\
\texttt{ranked\_mirna\_list} & Scores an ordered miRNA list after miRBase normalization. \\
\texttt{ranked\_protein\_site\_list} & Scores ordered protein+residue-site records, optionally including a case-insensitive site label. \\
\texttt{ranked\_residue\_site\_list} & Scores an ordered residue-site list after amino-acid and PTM-prefix normalization. \\
\texttt{ranked\_taxa\_list} & Scores an ordered list of taxa after NCBI Taxonomy normalization. \\
\texttt{record\_set} & Grades a variable-length list of objects using a per-field schema. \\
\texttt{relative\_numeric\_tolerance} & Compares one numeric value using error relative to the expected value. \\
\texttt{residue\_site\_set} & Scores an unordered set of residue-site strings after amino-acid and PTM-prefix normalization. \\
\texttt{scored\_entity\_set} & Scores unordered entity-score rows where the entity identity gates the row and a single numeric score receives fractional credit. \\
\texttt{scored\_interval\_set} & Scores an unordered set of genomic intervals, each carrying a numeric evidence field. \\
\texttt{spectral\_peak\_set} & Scores unordered spectral peak records with assignment, position, and intensity. \\
\texttt{string\_match} & Compares one scalar string using a selected literal, biological entity, amino-acid, or residue-site matching mode. \\
\texttt{string\_set} & Scores an unordered set of strings using a selected literal or semantic string matcher. \\
\texttt{struct\_set} & Scores unordered records made only of named string/entity fields. \\
\texttt{taxa\_set} & Scores an unordered set of taxa using NCBI Taxonomy synonym resolution. \\
\texttt{taxon\_pair\_set} & Scores unordered directed pair records whose endpoints are taxa. \\
\texttt{variant\_call} & Scores one small variant call with chromosome, position, reference allele, and alternate allele fields. \\
\texttt{variant\_set} & Scores an unordered set of small variant calls. \\
\texttt{weighted\_entity\_pair\_set} & Grades a set of network, interaction, regulatory, or association edges where each edge has two named endpoints (a, b) and a numeric weight, confidence, correlation, or effect size. \\
\end{longtable}

\section{Identifier databases}
\label{app:identifier-databases}

\begin{table}[H]
\centering
\caption{\textbf{Identifier databases that back OmicsBench's synonym-aware graders.}}
\label{tab:identifier-databases}
\small
\begin{tabular}{@{}p{2.6cm} p{\dimexpr\textwidth-2.6cm-3.2cm-4\tabcolsep\relax} p{3.2cm}@{}}
\toprule
Database & Description & Reference \\
\midrule
NCBI Gene & Matches between different gene nomenclature systems. & \citealt{brown_gene_2015} \\
NCBI Taxonomy & Matches between different organism naming systems. & \citealt{schoch_ncbi_2020} \\
HMDB & Matches between different metabolite naming systems. & \citealt{wishart_hmdb_2022} \\
miRBase & Matches between different miRNA naming systems. & \citealt{kozomara_mirbase_2019} \\
CARD & Matches between different antimicrobial resistance gene naming systems. & \citealt{alcock_card_2023} \\
\bottomrule
\end{tabular}
\end{table}

\section{A priori failure mode annotation}
\label{app:apriori-modes}

For each failed attempt, the annotating LLM judge received the subtask prompt, the grader configuration with the answer key, the submitted answer, the score and its sub-scores, and the agent's transcript for that turn, with model and harness names removed.
It could assign at most five failure modes to an attempt, each backed by a verbatim quote from the transcript.

To measure the LLM judge's reproducibility, we annotated a separate validation set of 100 failed attempts twice with the Opus 5 LLM judge, shuffling the attempts into different batches each time.
With each pairing of an attempt with one of the 13 failure modes treated as a yes-or-no decision on whether the mode was assigned, the two passes agreed with a Cohen's kappa of 0.76 over those decisions.

\begin{longtable}{@{}>{\small\centering\arraybackslash}p{1.0cm} >{\small\raggedright\arraybackslash}p{3.4cm} >{\small}p{\dimexpr\textwidth-4.4cm-4\tabcolsep\relax}@{}}
\caption{\textbf{A priori failure modes used to annotate failed attempts.}}\label{tab:apriori-modes} \\
\toprule
 & Failure mode & Definition \\
\midrule
\endfirsthead
 & Failure mode & Definition \\
\midrule
\endhead
\bottomrule
\endlastfoot
\swatch{fdae6b} & Environment setup failure & The agent could not get required tools, packages, references, or paths into a usable state, and the unresolved setup problem prevented or materially blocked the intended analysis. \\
\addlinespace[3pt]
\swatch{a1d99b} & Input misinterpretation & The agent misread the staged inputs, confused sample identities, or applied an analysis intended for a different data type. Applies only when the data themselves were misread; a later analysis choice that diverged is Method misconfigured. \\
\addlinespace[3pt]
\swatch{74c476} & Wrong method substitution & The agent used a method materially different from the one the reference analysis requires, such as a different tool family, resource, or statistical approach, including hand-built stand-ins (keyword matching, reimplemented pipelines) for a standard resource it did not use. \\
\addlinespace[3pt]
\swatch{31a354} & Method misconfigured & The agent used a suitable method but chose parameters, contrasts, references, normalizations, or filters that materially diverge from the reference analysis. Not added when a more specific mode captures the same choice. \\
\addlinespace[3pt]
\swatch{c7e9c0} & Incomplete data & The agent analyzed fewer samples, conditions, or features than the subtask specified, so the analysis was incomplete even where the reduced analysis ran. \\
\addlinespace[3pt]
\swatch{9e9ac8} & Output format violation & The answer did not match the subtask's answer contract: wrong filename or location, identifier namespace, naming, units, granularity, key names, or list length. \\
\addlinespace[3pt]
\swatch{756bb1} & Synthetic or placeholder output & The agent wrote synthetic, dummy, or memory-derived values instead of deriving the output from the staged data, including hand-assembled canonical panels substituted for its own computed result, whether openly or to conceal an incomplete analysis. \\
\addlinespace[3pt]
\swatch{e6550d} & Premature termination & The agent stopped before writing a scorable answer because it exhausted a turn or time budget, or declared completion before the work was finished. A timed-out attempt with no answer file is this mode even when no transcript exists. \\
\addlinespace[3pt]
\swatch{fd8d3c} & Inefficient analysis & The agent spent its time budget unproductively, on a materially large detour into work the subtask did not ask for or on re-running completed steps, and the lost time plausibly cost points. \\
\addlinespace[3pt]
\swatch{fdd0a2} & Repetitive retry loop & The agent repeatedly retried substantially the same failing command or approach without a meaningful diagnosis or change of strategy. \\
\addlinespace[3pt]
\swatch{3182bd} & Answer set too broad & The subtask asked for a compact, discriminating set; the agent submitted everything clearing a permissive cut or a fixed top-N block of its ranking, so co-ranked extras diluted the set's precision. \\
\addlinespace[3pt]
\swatch{6baed6} & Answer set too strict & The agent stacked stringent or conjunctive cutoffs, agreement demands, or exclusion lists until true members of the expected set were removed before submission. \\
\addlinespace[3pt]
\swatch{9ecae1} & Selection criterion mismatch & The agent selected or ranked candidates by a measure other than the one the subtask stated, so entities satisfying the wrong criterion filled the answer. Distinguished from Method misconfigured by being about which quantity ranked the candidates, not how it was computed. \\
\end{longtable}

\section{Unsupervised failure mode discovery}
\label{app:emergent-modes}

The failure descriptions were embedded with \texttt{openai/text-embedding-3-large}, reduced to five dimensions with UMAP (cosine distance, 15 neighbors, minimum distance 0), and clustered with HDBSCAN using leaf cluster selection and a minimum cluster size of 60, which gave 15 clusters holding 54\% of failed attempts.
The naming LLM judge saw each cluster's eight most central descriptions, with three descriptions from its boundary for contrast.
We set three of these clusters aside because they captured no failure of the agents.
Two captured tool outages (221 and 158 attempts), and the third merged attempts that used the whole time budget without writing an answer with attempts from a single task that took gene coordinates from the wrong genome version (137 attempts).

\begin{longtable}{@{}>{\small\centering\arraybackslash}m{1.0cm} >{\small\raggedright\arraybackslash}m{4.3cm} >{\small}m{\dimexpr\textwidth-5.3cm-4\tabcolsep\relax}@{}}
\caption{\textbf{Emergent failure modes recovered by unsupervised discovery.} The modes are grouped into five families.}\label{tab:emergent-modes} \\
\toprule
 & Failure mode & Definition \\
\midrule
\endfirsthead
 & Failure mode & Definition \\
\midrule
\endhead
\bottomrule
\endlastfoot
\multicolumn{3}{@{}l}{\small\itshape Too much submitted} \\*[2pt]
\modedisc{fcae91}{1} & Same number of top genes taken from every group, so lists were mostly extras & The agent took the same fixed number of top-ranked genes from every group, even though the subtask asked for the few genes that set each group apart, so most of what it submitted was extras, and the score fell. \\
\addlinespace[3pt]
\modedisc{fb6a4a}{2} & Every gene passing its thresholds submitted instead of a short list & The agent submitted every gene that passed its thresholds instead of selecting the short list the subtask asked for, so the submission was mostly extras. \\
\addlinespace[3pt]
\modedisc{de2d26}{3} & Ranked by test score only, so ubiquitous genes displaced the markers & The agent ranked genes by the test score alone with no expression filter, which let genes expressed everywhere take the top places, so the specific markers the subtask asked for were pushed off the list. \\
\addlinespace[3pt]
\modedisc{a50f15}{4} & Genes picked by keyword match instead of a statistical test & The agent chose genes by matching keywords in their names or annotations rather than by the statistical test the subtask required, which pulled in unrelated genes that matched the words and left out significant genes that did not. \\
\addlinespace[5pt]
\multicolumn{3}{@{}l}{\small\itshape Too little submitted} \\*[2pt]
\modedisc{8c510a}{5} & Two datasets each over-filtered before intersecting, so almost nothing remained & The agent filtered each of two datasets strictly on its own before intersecting them, when the subtask defined the answer as the intersection at a stated cutoff, so almost nothing survived, and most of the expected set was missing. \\
\addlinespace[5pt]
\multicolumn{3}{@{}l}{\small\itshape The agent's own criterion in place of the subtask's} \\*[2pt]
\modedisc{fee391}{6} & Extra cutoffs added on top of the subtask's criterion, dropping real genes and admitting unrelated ones & The agent added extra cutoffs on fold change, detection, and significance on top of the single criterion the subtask used, which dropped expected genes that missed any one cut while unrelated genes that happened to pass them all filled the list, so the answer both missed real members and carried extras. \\
\addlinespace[3pt]
\modedisc{fec44f}{7} & Only a self-written candidate list tested, so genes outside it were never found & The agent tested only a list of candidate genes it wrote itself rather than the whole dataset, so any expected gene missing from its list could never be found. \\
\addlinespace[3pt]
\modedisc{fe9929}{8} & Right category, but its example genes picked by a shortcut & The agent identified the right category but chose its example genes by keyword matching or its own ranking instead of taking them from the analysis behind the category, so the example list barely overlapped the expected one. \\
\addlinespace[5pt]
\multicolumn{3}{@{}l}{\small\itshape Prior knowledge in place of the data} \\*[2pt]
\modedisc{f768a1}{9} & Marker genes from memory added to the computed list, counted as extras & The agent added well-known marker genes from memory to the list it had computed, which put genes the dataset did not support into the answer, so they counted as extras. \\
\addlinespace[3pt]
\modedisc{ae017e}{10} & Answer written from memory without opening the data & The agent wrote the answer from memory or from earlier turns without opening the data, which produced the generic textbook answer where the subtask asked for the result specific to this dataset, so members were wrong or swapped. \\
\addlinespace[5pt]
\multicolumn{3}{@{}l}{\small\itshape A different procedure substituted} \\*[2pt]
\modedisc{66c2a4}{11} & Uncorrected p-values or a permissive test, so too many genes called significant & The agent used uncorrected p-values or the more permissive version of a test when the expected count came from the corrected test, causing it to report several times too many significant genes. \\
\addlinespace[3pt]
\modedisc{238b45}{12} & Time-course patterns assigned by hand-written rules instead of clustering & The agent sorted genes by the pattern of their expression over time using rules it wrote by hand, such as ``rises then falls'', instead of clustering the measured curves, so genes whose curves did not fit its rules landed in the wrong group. \\
\end{longtable}

\section{Model and harness configurations}
\label{app:model-configurations}

\begin{table}[H]
\centering
\caption{\textbf{Models and harnesses evaluated in this study.} Reasoning effort was set to high for every model except MiniMax-M3 and Qwen3.7-Max, for which OpenRouter exposes no effort setting. The harness versions were OpenCode 1.18.11, Claude Code 2.1.222, and Codex 0.142.5.}
\label{tab:models}
\small
\begin{tblr}{
  width = \textwidth,
  colspec = {X[l] X[l]},
  rows = {valign=m},
  hline{1,Z} = {0.08em},
  hline{2} = {0.05em},
  rowsep = 2pt,
}
Model & Harnesses \\
Opus 5 & OpenCode, Claude Code \\
GPT-5.6 Sol & OpenCode, Codex \\
Gemini 3.1 Pro & OpenCode \\
GLM-5.2 & OpenCode \\
Grok 4.3 & OpenCode \\
Grok 4.6 & OpenCode \\
Kimi K3 & OpenCode \\
MiniMax-M3 & OpenCode \\
Muse-Spark 1.1 & OpenCode \\
Qwen3.7-Max & OpenCode \\
Qwen3.8-Max & OpenCode \\
DeepSeek-V4-Pro & OpenCode \\
\end{tblr}
\end{table}

\section{Runtime environment details}
\label{app:runtime-environment}

Every run took place in its own Docker container, built from one standard image shared by all model/harness combinations.
The container ran Debian 13 (Trixie), and the host machine ran Ubuntu 24.04 LTS (Noble Numbat).
Each container was given four processor cores, 12 GB of memory, and 20 GB of storage, with the task dataset mounted read-only.
Containers had no network access, so an agent could use only what the image and the task dataset provided.
The Dockerfile that defines the benchmark runtime ships with the runner (\appref{app:code-data-availability}).
Within an OmicsBench session, the time limit per turn ($t_{\max}$) was 1800 seconds, and in gated mode the gate threshold $\tau$ was set at 0.7 and the attempt limit $\theta$ at 3 (\tabref{tab:parameters}).

\section{Judge prompts}
\label{app:judge-prompts}

Every judge call below carries the same system prompt:

\begin{promptverb}
You are an expert reviewer of computational biology work. You read transcripts of automated agents doing omics analyses inside a container and you describe what they did with precision and without judgement words. You answer with a single JSON object and nothing else: no prose before or after it, no code fences.
\end{promptverb}

\subsection{Trajectory narrative and a priori failure mode annotation}
\label{app:prompt-annotate}

\begin{promptverb}
You are annotating attempts from a benchmark of agents doing real omics analyses. Each agent worked in a container with the data under /data/ and had to write an answer file under /workspace/answers/. A grader then scored the answer file against a hidden key (score 0 to 1). The agent never saw the key.

Below are <batch size> evidence bundles, one attempt each, from different questions. Each bundle shows: the node prompt the agent received, the grader configuration with the expected value, the attempt record with the submitted answer, the score and its sub-scores, and the agent's own turn of the transcript (its messages, tool calls and tool results, truncated where long). Model names and harness names have been removed; the tool syntax may differ between bundles and that is not something to classify on. A bundle marked EVIDENCE MISSING has no transcript (the attempt timed out); it shows workspace file timestamps and the agent log instead.

Your job has two parts per bundle. Work only from the evidence in that bundle.

PART 1, for EVERY bundle: an approach-and-outcome narrative, two to three sentences on what the agent tried and what happened with it. Sentence one is the route taken: which files it opened, how it normalized or filtered, which statistical method or tool, what checks it ran. The remaining sentence or two give the mechanism of the result: how the specific choices led to the specific answer that was submitted, compared with what the grader expected. Write it the same way whether the score is high or low: describe, do not judge. DO NOT use verdict vocabulary anywhere in the narrative: no "succeeded", "failed", "failure", "correct", "incorrect", "wrong", "right", "mistake", "error", "bad", "good", "properly", "improperly". Name the route and the consequence instead ("used the raw intensity columns rather than the normalized ones, so the enriched set was dominated by contaminant proteins"). Also give the step index in the transcript where the route diverged from what the grader needed, or null if the score is at or above the threshold or you cannot tell.

ALSO for every bundle, a ROUTE field: ONE sentence describing the analytical route in fully task-agnostic terms. Abstract away every task-specific entity: no gene, protein, or pathway names, no tissue, disease, organism, or cell-type words, no file names, no database names beyond their kind (say "the annotation database", "the expression matrix", "the target entities", "the two groups"). Keep only the shape of the work: what kind of data was loaded, how it was transformed or normalized, what was computed or tested, how candidates were selected or ranked, what was checked, and what kind of thing was submitted. Example: "loaded the expression matrix, normalized by library size, ran a two-group differential test, kept everything passing a significance cut, and submitted the full list without pruning." This field is used to group attempts by route across different tasks, so any task-identifying word defeats its purpose.

PART 2, ONLY for bundles marked SCORE BELOW THRESHOLD: failure-mode tags from the fixed vocabulary below. Assign between zero and 5 tags: every mode whose mechanism the evidence concretely shows contributed to the score falling below 0.7. Require concrete evidence for every tag; when no mode applies, return an empty tag list (that is a valid and informative answer, never force a tag). For each tag give: the mode name copied verbatim; one or two sentences saying how this mechanism shows in this attempt; a verbatim quote of at most 200 characters from the transcript or the submitted answer showing it; and your confidence as "high" (the transcript shows the mechanism directly), "medium" (the best reading of partial evidence) or "low" (a plausible reading among several). If the bundle has no transcript and the evidence cannot support any judgement, use the single tag "undeterminable from evidence" with confidence "low" and say in the narrative's last sentence what evidence is missing.

The vocabulary. These are the only mode names PART 2 may use:

<the a priori failure modes, each as "- <name>: <definition>">
- undeterminable from evidence: The bundle does not contain enough evidence to support any judgement; used alone, only for bundles without a transcript.

Rules that override everything else:
- Tag the agent's work, never the harness, the tool syntax, or the container.
- A tag needs concrete evidence in this bundle; no tag when none applies.
- Tags name mechanisms that lost points, not incidental oddities. An attempt can carry several tags when several mechanisms each show concretely.
- Use only evidence inside the bundle. Do not use your own domain knowledge to decide what the key "should" be beyond what the grader configuration shows.
- Every quote must appear verbatim in the bundle. Never quote the grader's expected value.

Answer with ONE JSON object of exactly this shape, and nothing else:

{
  "bundles": [
    {
      "key": "<bundle key>",
      "narrative": "<two to three sentences>",
      "route": "<one task-agnostic sentence>",
      "diverged_at_step": <integer or null>,
      "tags": [
        {
          "mode": "<mode name from the vocabulary>",
          "mechanism": "<one or two sentences>",
          "evidence_quote": "<verbatim, at most 200 characters>",
          "confidence": "<high | medium | low>"
        }
      ]
    }
  ]
}

For a bundle whose score is at or above the threshold, set "tags" to null. For a below-threshold bundle where no mode applies, set "tags" to []. Bundle keys in this batch: <bundle keys>.

<evidence bundles>
\end{promptverb}

\subsection{Task-agnostic failure descriptions}
\label{app:prompt-abstract}

\begin{promptverb}
You will read <batch size> short narratives. Each describes how an automated agent's analysis of an omics dataset led to the answer it submitted, and how that answer related to what the grader expected. For EACH narrative write ONE sentence, in fully task-agnostic terms, describing the MECHANISM by which the submitted answer diverged from what was expected.

Rules for the sentence:
- Abstract away every task-specific entity: no gene, protein, pathway, cell-type, tissue, disease, organism, chromosome or sample names, no file, column or database names, no task-specific numbers. Say "the target features", "the two groups", "the annotation column", "the expected set", "the reference group".
- Keep the SHAPE of what happened: which decision or step produced the divergence (a selection rule, a threshold, a reference group, a normalisation, a label format, a data subset, a stopping point, a fabricated value), and how the submitted answer differed from the expected one in kind (more items, fewer items, different items, wrong format, no items, partially matching).
- Be specific about the mechanism, not the category. Do NOT write category labels such as "answer set too broad", "method misconfigured", "wrong method", "incomplete data", "premature termination", "placeholder output", "format violation"; write what the agent did instead ("kept every feature passing a nominal cut and submitted the whole list where a compact discriminating set was expected").
- Describe, do not judge: no "wrong", "incorrect", "failed", "error", "mistake", "properly".
- Write the same way for every score level. If the narrative says nothing about the mechanism (for example only that time ran out), say exactly that in task-agnostic terms.
\end{promptverb}

\subsection{Cluster naming}
\label{app:prompt-name}

\begin{promptverb}
You are naming one cluster of sentences. Each sentence describes, in task-agnostic terms, the mechanism by which an automated agent's submitted answer on an omics analysis task diverged from what was expected. The sentences were grouped by embedding similarity, so they should share one mechanism. You are given the <count> sentences closest to the cluster centre and, for contrast, sentences near the cluster boundary; the name should fit the former better than the latter.

Rules:
- The name is 3 to 8 words naming the MECHANISM as a bioinformatician would describe a colleague's choice: what the agent did and, if it fits, how the answer differed in kind (e.g. "fixed top-N block, extras dilute the set", "candidate panel chosen by hand, misses items outside it", "answer written from memory, data never opened").
- Name the mechanism, not a judgement and not a category: no "failed", "wrong", "incorrect", "error", "bad"; do not reuse taxonomy labels such as "answer set too broad" or "method misconfigured".
- Use only what the sentences show.
- If the sentences do not share one mechanism, set "coherent" to false and name the largest coherent subset.
\end{promptverb}

\end{document}